\documentclass{l4dc2027}
\title[Learning regions of attraction for TFL]{Learning Regions of Attraction for Transverse Feedback Linearization under Controller Singularities}
\usepackage{times}
\usepackage{booktabs}
\usepackage{array}
\newcommand{\Oad}{\Omega_{\mathrm{ad}}}
\newcommand{\R}{\mathbb{R}}
\newcommand{\D}{\mathcal D}
\newcommand{\I}{\mathcal I}
\newcommand{\Tm}{\mathcal T}
\newcommand{\qex}{q_{\mathrm{ex}}}
\newcommand{\dist}{\operatorname{dist}}
\newcommand{\relu}{\operatorname{relu}}

\newtheorem{problem}{Problem}
\newtheorem{assumption}{Assumption}

\coltauthor{\Name{Abdullah Waqar Malik} \Email{amalik.bscs22seecs@seecs.edu.pk}\\ \addr School of Electrical Engineering and Computer Science, NUST, Islamabad, Pakistan \AND \Name{Hamza Tariq} \Email{ht299@njit.edu}\\ \addr New Jersey Institute of Technology, Newark, NJ, USA \AND \Name{Adeel Akhtar} \Email{adeel.akhtar@njit.edu}\\ \addr New Jersey Institute of Technology, Newark, NJ, USA}

\begin{document}

\maketitle

\vspace{-0.35in}
\begin{abstract}
\looseness -1 Transverse feedback linearization (TFL) achieves path following by exactly linearizing the error dynamics, but its controller is undefined on a singular set, such as zero speed for a unicycle or zero thrust for a quadrotor. The constrained region of attraction, the set of initial states from which the closed loop converges to the path without violating state or regularity constraints, determines where the controller can be safely activated. Exploiting the exact linear error dynamics, we show that this region is the largest forward invariant subset of the admissible set. We then decompose its state-constrained Zubov value function into an analytically known quadratic cost-to-go and a safety value function. We learn only the latter, by fitted value iteration without infinite-horizon trajectory labels. The resulting estimate combines an inner layer verified by interval branch and bound with a learned outer layer tested on samples. On a unicycle following a circle and a sinusoid and on a planar quadrotor, it covers on average 72.0\,\%, 83.4\,\% and 19.1\,\% of the simulated region of attraction. In comparison, the quadratic cost-to-go and two physics-informed neural network (PINN) solvers of Zubov's equation cover at most 11.3\,\%, 4.0\,\% and 0.0\,\% on average. Code and animations are available at \url{https://gradslab.github.io/roa_tfl}.
\end{abstract}


\section{Introduction}\label{sec:intro}
\looseness -1 Set stabilization and path following are central problems in control and robotics \citep{nielsen2004,akhtar2015,allawati2026}. Transverse feedback linearization (TFL) addresses both with a coordinate transformation and a feedback that inverts a decoupling matrix \citep{nielsen2006,nielsen2008,akhtar2011}. The dynamics transverse to the path then become linear and exponentially stable, while the tangential and internal dynamics govern the motion along the path. TFL controllers have been designed for unicycles \citep{nielsen2004}, car-like robots and cars with trailers \citep{akhtar2011,akhtar2015,akhtar2011thesis}, quadrotors with and without a payload or a failed rotor \citep{roza2012,akhtar2012,akhtar2013,akhtar2022,akhtar2023,allawati2026,allawati2026ftc}, and robotic arms, also when carried by a quadrotor \citep{hladio2013,tariq2026}.

\looseness -1 In many of these works, the decoupling matrix and the Jacobian of the coordinate transformation are shown to be nonsingular on the path following manifold. By continuity and the inverse function theorem, they remain nonsingular on a neighbourhood of it, so the convergence guarantees are local \citep{nielsen2008,akhtar2015}. Beyond this neighbourhood, the domain of validity of TFL is limited by topological obstructions to globally regular path-following coordinates for closed paths. It is also limited by system-dependent singularities of the decoupling matrix, such as zero speed for the dynamically extended unicycle and zero thrust for the quadrotor \citep{nielsen2006,akhtar2015,allawati2026}. Consequently, a trajectory that starts far from the path may reach these singularities even though its transverse error decays exponentially while the controller is defined. We therefore seek the constrained region of attraction, that is, the set of initial states from which the closed-loop trajectories converge to the path while the controller remains well defined and all state constraints hold for all future time.

\looseness -1 The Zubov value function characterizes regions of attraction \citep{camilli2001}, and with an obstacle term for the state constraints it is finite precisely on the constrained region of attraction \citep{gruene2015}. The difficulty is computing it. Grid-based solvers of the associated Hamilton-Jacobi equation suffer from the curse of dimensionality \citep{mitchell2005}, and sum-of-squares methods \citep{xue2020} require polynomial dynamics, whereas TFL closed loops are rational, trigonometric and singular. Neural solvers of Zubov equations treat equilibria or, in discrete time, compact invariant sets \citep{liu2025,kang2023,mengliu2026,serry2025,serry2026,meng2024}. Most are trained on labels from trajectories simulated until convergence, and none handles a path-following target set together with controller singularities. On the TFL closed loops of this paper, we measure coverage against the region of attraction found by simulating a set of reference states, which we call the simulated region of attraction (Section~\ref{sec:exp}). Two such solvers \citep{liu2025,mengliu2026}, both physics-informed neural networks (PINNs), cover on average at most 11.3\,\%, 4.0\,\% and 0.0\,\% of it. To the best of our knowledge, no existing method learns the region of attraction of a TFL closed loop. The closest works on feedback linearization and path following \citep{oishi2006,faulwasser2016} do not estimate it for a given controller (Appendix~\ref{app:related}).

\looseness -1 Our key observation is that existing Zubov solvers do not exploit the exact linear error dynamics of TFL. On the admissible set, where the controller is well defined, the coordinate transformation is regular and the state constraints hold, these dynamics make the constrained region of attraction the largest forward invariant subset. They also split the state-constrained Zubov value function exactly into a known quadratic cost-to-go and a safety value function that accounts for the singularities and the constraints. As a result, only the safety value function has to be learned, from short trajectory segments and without infinite-horizon trajectory labels. These results hold for arbitrary relative degrees, an optionally regulated rate of progress and internal dynamics that need not be stable. Beyond TFL, the same algorithm applies to closed loops with an exponentially stable equilibrium, with the quadratic form of the linearization in place of the exact cost-to-go. The decomposition is then no longer exact, yet the estimates cover between 86.5\,\% and 98.9\,\% of the simulated region of attraction on one seed of five benchmarks (Appendix~\ref{app:bench}). Our main contributions are as follows.

\vspace{-0.1in}
\begin{itemize}\setlength{\itemsep}{0pt}\setlength{\parskip}{6pt}\setlength{\topsep}{3pt}
\item We characterize the constrained region of attraction of a TFL closed loop as the largest forward invariant subset of the admissible set (Theorem~\ref{thm:gz}). We also decompose its Zubov value function into a known quadratic cost-to-go and a safety value function (Lemma~\ref{lem:split}).
\item We propose an algorithm that learns only the safety value function, by fitted value iteration without infinite-horizon trajectory labels (Algorithm~\ref{alg:method}).
\item We construct a two-layer estimate with a verified inner layer and an outer layer tested on samples (Theorem~\ref{thm:main}), with statistical bounds on what the tests can miss (Section~\ref{sec:guarantees}). On a unicycle and a planar quadrotor, it covers more of the simulated region of attraction than the quadratic cost-to-go and two PINN solvers of Zubov's equation (Table~\ref{tab:main}).
\end{itemize}

\vspace{-0.1in}
Section~\ref{sec:problem} formulates the problem, Section~\ref{sec:method} presents the method, Section~\ref{sec:guarantees} states its guarantees, and Section~\ref{sec:exp} reports the simulations. The appendices give the proofs and further details.

\vspace{-0.04in}
\paragraph{Notation.}\looseness -1 For a function $\varphi$ and a vector field $f$, $L_f\varphi=d\varphi(f)$ is the Lie derivative, written $\langle\nabla\varphi,f\rangle$ in coordinates. We write $y(t;x)$ for the solution of the closed loop from $x$ in physical time, and $\dist$ for the distance of a fixed Riemannian metric on the state space, with $\dist(x,M)=\inf_{x'\in M}\dist(x,x')$. Moreover, $\relu(\cdot)=\max\{\cdot,0\}$, $B^\circ$ is the interior of a box $B$, and $\{\varphi<c\}$ is a sublevel set of $\varphi$.

\vspace{-0.15in}
\section{Preliminaries and Problem Formulation}\label{sec:problem}
\subsection{Transverse feedback linearization}\label{sec:tfl}
\looseness -1 We follow the geometric formulation of TFL of \citet{nielsen2008}, and its use for path following with a regulated rate of progress of \citet{akhtar2015}, and we refer to these works for details. Consider a smooth control-affine system
\begin{equation}\label{eq:sys}
\dot x=f(x)+\sum_{j=1}^m g_j(x)\,u_j
\end{equation}
on an $n$-dimensional smooth manifold $X$, where $u=(u_1,\dots,u_m)\in\R^m$ and $f,g_1,\dots,g_m$ are smooth vector fields on $X$. Let $p:X\to Y$ be a smooth position map, where $Y$ is $\R^{\ell+1}$, $\ell\ge1$, or its quotient by periods in some coordinates, with the metric induced by the Euclidean metric. Let $U\subset Y$ be open and let $\alpha=(\alpha_1,\dots,\alpha_\ell):U\to\R^\ell$ be smooth, defining a regular path $\Gamma=\{\alpha=0\}\subset U$ with $\operatorname{rank}d\alpha=\ell$ on $\Gamma$. Let $\varpi:U\to\mathcal P$, with $\mathcal P=\R$ or $\mathbb S^1$, be a smooth progress map with $d(\varpi|_\Gamma)\ne0$ on $\Gamma$. On $X_U=p^{-1}(U)$, we write $\alpha$ and $\varpi$ for $\alpha\circ p$ and $\varpi\circ p$.

\looseness -1 The controller regulates the transverse outputs $\alpha_1,\dots,\alpha_\ell$ and, optionally, the rate of progress along $\Gamma$. Let $r_1,\dots,r_\ell\ge1$ and, when the rate of progress is regulated, $r_0\ge2$ be integers such that, on $X_U$ and for $j=1,\dots,m$, $L_{g_j}L_f^\iota\alpha_i=0$ for $0\le\iota<r_i-1$ and $L_{g_j}L_f^\iota\varpi=0$ for $0\le\iota<r_0-1$. The transverse coordinates are $\xi=(\xi^1,\dots,\xi^\ell)\in\R^r$ with $\xi^i=(\alpha_i,L_f\alpha_i,\dots,L_f^{r_i-1}\alpha_i)$ and $r=r_1+\dots+r_\ell$, and the tangential coordinates are $\eta=(\varpi,L_f\varpi,\dots,L_f^{r_0-1}\varpi)$. If the rate of progress is regulated, let $k=\ell+1$ and let the coordinate transformation be $\Tm=(\eta,\xi)$. Otherwise, let $k=\ell$ and $\Tm=\xi$. In both cases we assume $k\le m$. The decoupling matrix $D:X_U\to\R^{k\times m}$ has the rows $(L_{g_j}L_f^{r_0-1}\varpi)_{j=1}^m$, in the first case only, and $(L_{g_j}L_f^{r_i-1}\alpha_i)_{j=1}^m$ for $i=1,\dots,\ell$. Wherever $D$ has full row rank, the regulated outputs have vector relative degree $(r_0,r_1,\dots,r_\ell)$, or $(r_1,\dots,r_\ell)$, and the TFL controller is
\begin{equation}\label{eq:tfl}
u=D(x)^\dagger\left(-\begin{bmatrix}L_f^{r_0}\varpi\\ L_f^{r_1}\alpha_1\\ \vdots\\ L_f^{r_\ell}\alpha_\ell\end{bmatrix}+\begin{bmatrix}\nu_\parallel\\ \nu_{\perp,1}\\ \vdots\\ \nu_{\perp,\ell}\end{bmatrix}\right),\qquad \nu_{\perp,i}=\Lambda_i^\top\xi^i,\qquad \nu_\parallel=\Lambda_\eta^\top(\eta_2-\bar\eta,\eta_3,\dots,\eta_{r_0}),
\end{equation}
where the first row of each vector is present only when the rate of progress is regulated, $D^\dagger=D^\top(DD^\top)^{-1}$ is the right inverse of $D$ (equal to $D^{-1}$ when $k=m$), and $\bar\eta\in\R$ is the reference rate of progress. The error coordinates are $z=(\xi,\eta_2-\bar\eta,\eta_3,\dots,\eta_{r_0})$ when the rate of progress is regulated and $z=\xi$ otherwise, with dimension $n_z=r+r_0-1$ or $n_z=r$. The gains $\Lambda_i\in\R^{r_i}$ and $\Lambda_\eta\in\R^{r_0-1}$ are chosen such that $\dot z=Az$ with $A$ Hurwitz and block diagonal. The remaining $n-n_z$ dimensions of the state carry the internal dynamics of the closed loop, which contain the motion along $\Gamma$ \citep{nielsen2006}. They may be unstable, and our results do not assume their stability.

\paragraph{Path following manifold.} As in \citet{nielsen2006} and \citet{akhtar2015}, the path following manifold $\Gamma^\star$ is the maximal controlled invariant subset of the lift $\tilde\Gamma=\{x\in X_U:\alpha(x)=0\}$ of the path. In other words, it is the set of states from which a control signal can keep the position on $\Gamma$ for all future time. Wherever the rows of $D$ that belong to $\alpha_1,\dots,\alpha_\ell$ are linearly independent, $\Gamma^\star$ is the embedded submanifold $\{\xi=0\}$ of dimension $n-r$ (Appendix~\ref{app:systems}).

\paragraph{Examples.} Our experiments use two systems that regulate the rate of progress, both with $m=k=2$. The dynamically extended unicycle has $n=4$, $r_1=r_0=2$ and $n_z=3$, and it follows a unit circle and a sinusoid. The planar quadrotor with the thrust as a controller state has $n=6$, $r_1=r_0=3$ and $n_z=5$, and it follows a vertical unit circle. Appendix~\ref{app:systems} gives their dynamics, gains, thresholds and state constraints.

\vspace{-0.075in}
\subsection{The admissible set}\label{sec:oad}
\looseness -1 Let $B\subset X$ be a compact box in a fixed chart, with periodic coordinates identified, whose bounds are the state constraints. In this chart, let $J_\Tm$ be the Jacobian of $\Tm$, let $\Delta_\Tm=\sqrt{\det(J_\Tm J_\Tm^\top)}$, and let $\Delta_D=\sqrt{\det(DD^\top)}$, which equals $|\det D|$ when $k=m$. The controller \eqref{eq:tfl} is undefined where $D$ loses rank, the coordinate transformation $\Tm$ loses rank where $\Delta_\Tm=0$, and the progress map may lose smoothness where its gradient is unbounded, as at the centre of a circular path. We call the union of these sets the singular set. Given thresholds $\delta_D,\delta_\Tm,\delta_\varpi>0$, the admissible set is
\begin{equation}\label{eq:oad}
\Oad=\{x\in X_U\cap B^\circ:\ \Delta_D(x)>\delta_D,\ \Delta_\Tm(x)>\delta_\Tm,\ \|\nabla\varpi(p(x))\|<\delta_\varpi^{-1}\},
\end{equation}
where the last inequality is imposed only when the rate of progress is regulated. The closure $\overline{\Oad}$ is compact because $B$ is. We assume $\overline{\Oad}\subset X_U$ and call the inequalities in \eqref{eq:oad} the regularity constraints. Their positive thresholds keep $\overline{\Oad}$ off the singular set, so the closed-loop vector field $f_{\mathrm{cl}}$ of \eqref{eq:sys} and \eqref{eq:tfl} is smooth on a neighbourhood of $\overline{\Oad}$. Since $u$ scales with $1/\Delta_D$, $f_{\mathrm{cl}}$ grows large near the singular set, so we integrate the normalized field $\bar f=f_{\mathrm{cl}}/a$, where $a=\sqrt{1+\|Sf_{\mathrm{cl}}\|^2}$ and $S$ is the diagonal matrix of inverse half-widths of $B$. This change of time scale preserves orbits, forward invariance and convergence within $\Oad$, and all step lengths in training and in the tests are measured in the time of $\bar f$, which we call normalized time.

\vspace{-0.075in}
\subsection{Target set and constrained region of attraction}\label{sec:roa}
The target set is $M=\{x\in\overline{\Oad}:z(x)=0\}$. Since $\Delta_D\ge\delta_D>0$ on $\overline{\Oad}$, $D$ has full row rank there, so $M\subset\Gamma^\star$. Without regulation of the rate of progress, $M=\Gamma^\star\cap\overline{\Oad}$. With it, $M$ is the part of $\Gamma^\star\cap\overline{\Oad}$ on which the path is traversed at the reference rate. We make the following assumption.

\vspace{-0.075in}
\begin{assumption}\label{ass:main}
(a) The closed-loop vector field $f_{\mathrm{cl}}$ and the error map $z$ are smooth on a neighbourhood of $\overline{\Oad}$. (b) $L_{f_{\mathrm{cl}}}z(x)=Az(x)$ for all $x\in\Oad$, with $A$ Hurwitz. (c) $\varnothing\ne M\subset\Oad$, so $z\ne0$ on $\partial\Oad$.
\end{assumption}

\vspace{-0.075in}
Parts (a) and (b) follow from Sections~\ref{sec:tfl} and~\ref{sec:oad}. Part (c) keeps the motion along the path inside the state constraints and off the singular set, and Appendix~\ref{app:systems} checks it for our examples. The set $M$ is compact and forward invariant (Appendix~\ref{app:proofs}), and it may be disconnected, as for the unicycle, which can follow the path driving forward or in reverse (Appendix~\ref{app:systems}). The constrained region of attraction of $M$ is
\begin{equation}\label{eq:D}
\D(M)=\bigl\{x\in\Oad:\ y(t;x)\in\Oad\ \text{for all }t\ge0\ \text{ and }\ \lim_{t\to\infty}\dist(y(t;x),M)=0\bigr\},
\end{equation}
By Theorem~\ref{thm:gz}, a trajectory from a state of $\Oad\setminus\D(M)$ leaves $\Oad$, by violating a regularity constraint or a state constraint. Deciding membership in $\D(M)$ directly thus requires simulation.

\vspace{-0.075in}
\begin{problem}\label{prob:main}
Given $f_{\mathrm{cl}}$, $M$ and $\Oad$ satisfying Assumption~\ref{ass:main}, compute a set $\Omega$ that contains a neighbourhood of $M$, is described by sublevel sets of explicit functions, and lies in $\D(M)$ up to a set of bounded measure, with a stated confidence. The set must be obtained without simulating trajectories until they converge, and it should cover as large a share of $\D(M)$ as possible.
\end{problem}

\vspace{-0.2in}
\section{Methodology}\label{sec:method}
We address Problem~\ref{prob:main} with a two-layer estimate. The inner layer $\I$ is a sublevel set of a known quadratic function of the error coordinates, and its inclusion in $\D(M)$ is verified by interval branch and bound. The outer layer is a sublevel set of a learned approximation of the state-constrained Zubov value function, and its inclusion is tested on samples. The estimate $\Omega$ is the union of the two layers. This section describes its steps, which Algorithm~\ref{alg:method} collects at the end, and Section~\ref{sec:guarantees} states the guarantees of each layer.

\vspace{-0.15in}
\subsection{Decomposition of the value function}
Let $\zeta=S_zz$, where $S_z\in\R^{n_z\times n_z}$ is a positive diagonal scaling matrix. Since $\tilde A=S_zAS_z^{-1}$ is Hurwitz, for every symmetric $Q\succ0$ the Lyapunov equation $\tilde A^\top P+P\tilde A=-Q$ has a unique symmetric solution $P\succ0$. The quadratic cost-to-go is $\qex(x)=\zeta(x)^\top P\zeta(x)$, and the running cost is $\mathcal L(x)=\gamma\,\zeta(x)^\top Q\zeta(x)$ with a cost scale $\gamma>0$. Appendix~\ref{app:systems} gives $S_z$, $Q$ and $\gamma$ for our examples.

Let $d_i$, $i=1,\dots,N_c$, be the distances to the regularity constraints of \eqref{eq:oad} and to the nonperiodic faces of $B$. Each $d_i$ is locally Lipschitz on a neighbourhood of $\overline{\Oad}$ and positive on $\Oad$, and at every point of $\partial\Oad$ at least one $d_i$ vanishes. The obstacle function $h:X_U\to[0,\infty]$ is
\begin{equation}\label{eq:h}
h(x)=\begin{cases}\displaystyle\sum_{i=1}^{N_c}\relu\big(b_i/d_i(x)-1\big)^2, & x\in\Oad,\\ +\infty, & x\notin\Oad,\end{cases}
\end{equation}
where the constants $b_i$ satisfy $0<b_i<\min_{x\in M}d_i(x)$. Hence $h$ is locally Lipschitz on $\Oad$, vanishes on a neighbourhood of $M$, and tends to infinity at $\partial\Oad$. For a state whose trajectory exists and stays in $\Oad$ for all $t\ge0$, the state-constrained Zubov value function of \citet{gruene2015} and its bounded transform are
\begin{equation}\label{eq:V}
V(x)=\sup_{t\ge0}\Big[\int_0^t\mathcal L(y(s;x))\,ds+h(y(t;x))\Big],\qquad v(x)=1-e^{-V(x)},
\end{equation}
with integration in physical time. For every other state, $V(x)=+\infty$ and $v(x)=1$. Theorem~\ref{thm:gz} in Section~\ref{sec:guarantees} shows that $\D(M)=\{x\in\Oad:V(x)<\infty\}=\{x\in\Oad:v(x)<1\}$.

\begin{lemma}\label{lem:split}
Let Assumption~\ref{ass:main} hold. Then $L_{f_{\mathrm{cl}}}\qex(x)=-\zeta(x)^\top Q\zeta(x)$ for all $x\in\Oad$. Consequently, for every $x\in\Oad$ and $t\ge0$ such that $y(s;x)\in\Oad$ for all $s\in[0,t]$, $\int_0^t\mathcal L(y(s;x))\,ds=\gamma\big(\qex(x)-\qex(y(t;x))\big)$. Define the margin function $\psi:\Oad\to[0,\infty)$ by $\psi=\relu(h-\gamma\qex)$. For every $x\in\Oad$,
\begin{equation}\label{eq:split}
V(x)=\gamma\qex(x)+\Phi(x),\qquad \Phi(x)=\sup_{t\ge0}\psi(y(t;x)),
\end{equation}
where $\Phi(x)=+\infty$ if the trajectory of $x$ leaves $\Oad$. In particular, $V\ge\gamma\qex$ and $\Phi\ge\psi\ge0$ on $\Oad$. Moreover, for every $\tau>0$ and every $x\in\Oad$ such that $y(s;x)\in\Oad$ for all $s\in[0,\tau]$,
\begin{equation}\label{eq:dp}
\Phi(x)=\max\Big\{\max_{0\le s\le\tau}\psi(y(s;x)),\ \Phi(y(\tau;x))\Big\}.
\end{equation}
\end{lemma}
\looseness -1 The proof is in Appendix~\ref{app:proofs}. Identity \eqref{eq:dp} has the form of the self-consistency equation of an undiscounted safety value function \citep{yu2022}, so we call $\Phi$ the safety value function. Since $\gamma\qex$ is known in closed form, only $\Phi$ has to be learned, and the quadratic term supplies the strict decrease of $V$ along trajectories where $\Phi$ is constant. Moreover, the bound $V\ge\gamma\qex$ holds for every approximation of the form $\gamma\qex$ plus a nonnegative function, which gives the inclusion of Theorem~\ref{thm:main}(ii). When the inner layer is not negligible, we add a shaping cost to $\mathcal L$, to steepen $v$ just outside the inner layer. Its integral is not known in closed form, so $\Phi$ absorbs it, and the decomposition, the training and the guarantees extend to it (Appendix~\ref{app:config}). For brevity, this section presents the method without it.

\vspace{-0.075in}
\subsection{Neural candidate}\label{sec:cand}
We approximate $\Phi$ on $\Oad$ in the form
\begin{equation}\label{eq:ansatz}
\begin{aligned}
\Phi_\theta&=\max\{\psi,\Psi_\theta\},\qquad V_\theta=\gamma\qex+\Phi_\theta,\qquad v_\theta=1-e^{-V_\theta},\\
\Psi_\theta(x)&=\tanh\!\big(\qex(x)/q_0\big)\,\tfrac1\beta\,\mathrm{softplus}\big(\beta\,\mathcal N_\theta(F(x))\big),
\end{aligned}
\end{equation}
with $V_\theta=+\infty$ and $v_\theta=1$ outside $\Oad$, and we call $v_\theta$ the candidate. Here $\mathcal N_\theta$ is a multilayer perceptron with weights $\theta$ and $\beta=10$. Its features $F(x)$ are built from the state, the error coordinates and the distances to the singular set (Appendix~\ref{app:config}). The constant $q_0>0$ is the first percentile of $\qex$ over the collocation points, the admissible samples on which the network is trained (Appendix~\ref{app:config}), and the factor $\tanh(\qex/q_0)$ gives $\Psi_\theta=0$ on $M$. Like the exact solution, $V_\theta$ is $\gamma\qex$ plus a function that is at least $\psi$. Hence $V_\theta\ge\max\{h,\gamma\qex\}$ on $\Oad$, and every sublevel set $\{V_\theta<c\}$ with $c<\infty$ lies in $\{h<c\}\subset\Oad$. These sublevel sets therefore satisfy the regularity constraints and the state constraints by construction, although their forward invariance is not automatic.

\vspace{-0.075in}
\subsection{Training and level selection}\label{sec:train}
\paragraph{Fitted value iteration.} Training solves \eqref{eq:dp} by regression. The update follows the semi-Lagrangian scheme (5.1) of \citet{gruene2015}, with a target network as in reach-avoid reinforcement learning \citep{hsu2021}. Let $\bar y(s;x)$ denote the solution of $\dot x=\bar f(x)$. Since \eqref{eq:dp} concerns the orbit only, it holds in normalized time and for every step $\tau>0$. For each collocation point $x$, we integrate $\bar f$ by the fourth-order Runge-Kutta method (RK4) over a step $\tau$ with substeps of length $0.025$. We then store the end point $x_\tau$ and the running maximum $R_\tau(x)=\max_{0\le s\le\tau}\psi(\bar y(s;x))$ over the substep end points, including $s=0$, with $R_\tau(x)=+\infty$ if a substep leaves $\Oad$. These quantities do not depend on the network, so they are computed once. Training keeps an exponential moving average of the weights. With a copy $\bar\theta$ of these averaged weights, refreshed after every epoch, the target is
\begin{equation}\label{eq:target}
\mathcal U(x)=\exp\!\big(-\max\{R_\tau(x),\ \Phi_{\bar\theta}(x_\tau)\}\big)\in[0,1],
\end{equation}
which is bounded even where the exact value is infinite and equals zero when the step leaves $\Oad$. Since $R_\tau(x)\ge\psi(x)$, the loss is $(e^{-\Psi_\theta(x)}-\mathcal U(x))^2$ where $\mathcal U(x)<e^{-\psi(x)}$, and elsewhere the one-sided loss $\relu(\mathcal U(x)-e^{-\Psi_\theta(x)})^2$ only keeps $\Psi_\theta$ below $\psi$, so that zero loss implies $e^{-\Phi_\theta(x)}=\mathcal U(x)$. Finally, for a point whose step leaves $\Oad$, this loss and the obstacle penalty below carry the factor $1+w_{\mathrm{ex}}$ with an exit weight $w_{\mathrm{ex}}\ge0$.

\vspace{-0.075in}
\paragraph{Step setting.}\looseness -1 Unrolling \eqref{eq:target} over successive steps shows that its only continuous fixed point that vanishes on $M$ is $\sup_{t\ge0}\psi(y(t;x))$ for every step $\tau$, which is $\Phi$ because the decomposition is exact (Appendix~\ref{app:config}). A long step is therefore admissible on a TFL closed loop, and it carries the value of its end point over a long arc of the orbit in each update. Since the best step within a training budget still depends on the system, the method does not fix it in advance. It trains seed 0 with a long-step setting ($\tau=2$, $\gamma=1/q_{90}$, $w_{\mathrm{ex}}=10$) and with a short-step setting ($\tau=0.2$, $\gamma=3/q_{90}$, $w_{\mathrm{ex}}=0$), where $q_{90}$ is the 90th percentile of $\qex$ over admissible samples. It then keeps, for all seeds, the setting whose estimate holds the larger share of box samples after the level selection and the final test described below, before fine-tuning. This share is the score by which level selection ranks snapshots, and it uses no reference labels.

\vspace{-0.075in}
\paragraph{Penalties.} \looseness -1 In addition to the regression loss, three penalties encode properties of the exact solution. The finite-step penalty asks $V_\theta$ to decrease over steps of unit length in normalized time that stay in $\Oad$, the decrease penalty asks for $L_{f_{\mathrm{cl}}}\Phi_\theta\le0$ where $\Psi_\theta>\psi$, and the obstacle penalty asks for $\Psi_\theta>\psi$ where $L_{f_{\mathrm{cl}}}\psi>0$. Except on the quadrotor, every term carries an exponential loss weight, which keeps the gradient at the levels near one where the outer level is decided. In addition, the states whose unit step violates the decrease most join the data during training (Appendix~\ref{app:config}).

\vspace{-0.075in}
\paragraph{Level selection.} The inner layer is $\I=\{x\in B^\circ\cap X_U:\gamma\qex(x)<c_I\}$, with $c_I$ one percent below the largest tested level that passes the interval verification of Section~\ref{sec:guarantees}, and $c_1=1-e^{-c_I}$. A state $x\in\Oad$ satisfies the finite-step decrease condition for a function $\tilde v:\Oad\to[0,1)$ if, for some horizon $H\in\{1,2,4,8\}$, $\bar y(s;x)\in\Oad$ for all $s\in[0,H]$ and $\tilde v(\bar y(H;x))<\tilde v(x)$. In level selection, $\tilde v=v_\theta$. A state with $v_\theta(x)\ge c_1$ that fails this condition is a counterexample. The condition is tested numerically, with the exemptions of Appendix~\ref{app:config}. The outer level $c_2$ is the lowest value of $v_\theta$ among the sampled counterexamples, or one if none is found, and we call $\{c_1\le v_\theta<c_2\}$ the annulus. Training saves a snapshot of the network every five epochs. Screening sets $c_2$ for every snapshot from counterexamples sought on samples of the box, of its faces and next to the singular set, and ranks the snapshots by the share of box samples in $\I\cup\{v_\theta<c_2\}$. The three best are then retested on fresh samples, and the best one after the retest is kept (Appendix~\ref{app:config}).

\vspace{-0.075in}
\paragraph{Counterexample-guided fine-tuning.} Since the outer level is the lowest value among the counterexamples, a handful of states determines it. Each of four rounds of fine-tuning therefore searches fresh samples for counterexamples with $v_\theta(x)\le c_2+0.05$ and runs 300 Adam steps on them, with a decrease margin and an anchor term that holds $\Psi_\theta$ elsewhere. The networks of these rounds and the trained network are screened like training snapshots. With the kept network and its level $\bar c_2$ fixed, the final test draws $N=10^6$ i.i.d.\ uniform samples of the box, with a seed disjoint from those of all preceding stages. The final level $c_2$ is then the minimum of $\bar c_2$ and the values of $v_\theta$ at the counterexamples that this test finds. If the kept network is the trained one, the counterexamples of its earlier final test count as well. Algorithm~\ref{alg:method} summarizes the method.

\vspace{-0.15in}
\section{Guarantees}\label{sec:guarantees}
This section characterizes $\D(M)$, gives sufficient conditions for the inclusion of both layers of the estimate $\Omega=\I\cup\{v_\theta<c_2\}$ in $\D(M)$, and states what the tests of the outer layer guarantee. The proofs are in Appendix~\ref{app:proofs}.

\begin{theorem}\label{thm:gz}
Let Assumption~\ref{ass:main} hold. A state $x\in\Oad$ belongs to $\D(M)$ if and only if its trajectory exists and remains in $\Oad$ for all $t\ge0$. Consequently, $\D(M)$ is the largest forward invariant subset of $\Oad$, and $\D(M)=\{x\in\Oad:V(x)<\infty\}=\{x\in\Oad:v(x)<1\}$.
\end{theorem}

The set equality is the conclusion of Theorem~3.1 of \citet{gruene2015}, here for a target set that may be disconnected and without stability assumptions on the internal dynamics. The next result gives sufficient conditions for the inclusion of both layers in $\D(M)$.

\begin{algorithm2e}[H]
\caption{Two-layer estimate of the constrained region of attraction of a TFL closed loop}\label{alg:method}
\small\DontPrintSemicolon
\KwIn{closed loop $f_{\mathrm{cl}}$, admissible set $\Oad$, quadratic cost-to-go $\qex$, obstacle function $h$, set $\mathcal S$ of step settings $(\tau,\gamma,w_{\mathrm{ex}})$}
\KwOut{estimate $\Omega=\I\cup\{v_\theta<c_2\}$}
\textbf{Inner layer:} set $c_I$ one percent below the largest tested level that passes the interval verification of Section~\ref{sec:guarantees}, and set $\I=\{\gamma\qex<c_I\}$, $c_1=1-e^{-c_I}$\;
\textbf{Step setting:} on seed 0, run the lines below up to the level selection and a final test for every setting in $\mathcal S$, and keep, for all seeds, the setting whose estimate holds the largest share of box samples\;
\textbf{Step data:} from every collocation point $x$, integrate $\bar f$ over the step $\tau$ and store the end point $x_\tau$ and the running maximum $R_\tau(x)$\;
\For{each epoch}{
freeze a copy $\bar\theta$ of the averaged weights and form the target $\mathcal U$ of \eqref{eq:target}\;
update $\theta$ on the regression loss of $\Phi_\theta=\max\{\psi,\Psi_\theta\}$ to $\mathcal U$, with the factor $1+w_{\mathrm{ex}}$ on points whose step leaves $\Oad$, plus the three penalties\;
add the states that violate the finite-step decrease most to the data\;
}
\textbf{Level selection:} for every snapshot, set $c_2$ to the lowest value of $v_\theta=1-e^{-(\gamma\qex+\Phi_\theta)}$ among the sampled counterexamples, and keep the snapshot whose estimate holds the largest share of box samples\;
\textbf{Fine-tuning:} for four rounds, search fresh samples for counterexamples with $v_\theta\le c_2+0.05$, train on them and repeat the level selection\;
\textbf{Final test:} draw $N=10^6$ i.i.d.\ samples and lower $c_2$ to the lowest value of $v_\theta$ among the counterexamples found\;
\end{algorithm2e}

\begin{theorem}\label{thm:main}
Let Assumption~\ref{ass:main} hold, let $c_I>0$, $\I=\{x\in B^\circ\cap X_U:\gamma\qex(x)<c_I\}$ and $c_1=1-e^{-c_I}$. (i) Suppose that every $x\in B\cap X_U$ with $\gamma\qex(x)<c_I$ satisfies the regularity constraints of \eqref{eq:oad}, and that at every such point on $\partial B$ the field $f_{\mathrm{cl}}$ points strictly into $B$ across every active face. Then $\I$ is forward invariant and $M\subset\I\subset\D(M)$. (ii) Let $\tilde v:\Oad\to[0,1)$ satisfy $-\ln(1-\tilde v)\ge\gamma\qex$, as $v_\theta$ does for every $\theta$. Then $\{\tilde v<c_1\}\subset\I$. (iii) Let (i) hold, let $\tilde v$ be as in (ii) and let $c_2\in[c_1,1]$. Suppose that every $x\in\Oad\setminus\I$ with $c_1\le\tilde v(x)<c_2$ satisfies the finite-step decrease condition for $\tilde v$. Then $\{\tilde v<c_2\}\subset\D(M)$.
\end{theorem}

\looseness -1 The inclusion $\{v_\theta<c_1\}\subset\I$ of part (ii) holds by construction for every candidate, and we check the hypotheses of part (i) by interval branch and bound \citep{moore2009} in floating-point arithmetic with outward rounding (Appendix~\ref{app:inner}).

\vspace{-0.075in}
\paragraph{Statistical bounds.} Two bounds quantify what the sample tests can miss (Appendix~\ref{app:proofs}). First, the final test fixes the network and the level $\bar c_2$ before it draws its $N$ samples, so a scenario bound applies to it (Proposition~\ref{prop:scen}). With probability at least $1-(1-\varepsilon)^N$, the counterexamples of the implemented test that remain in the annulus have measure at most $\varepsilon$. However, such a bound does not bound the measure of the states of the outer layer outside $\D(M)$ \citep{kenanian2019}. Second, the direct test of unsafe states therefore simulates $N$ fresh i.i.d.\ samples until they reach the inner layer and counts the $K$ samples of an estimate that fail to do so inside $\Oad$. With confidence $1-10^{-6}$, the states of the estimate outside $\D(M)$ then have measure at most the one-sided Clopper-Pearson bound $\varepsilon_K$ (Proposition~\ref{prop:unsafe}), which is about $\ln(10^6)/N$ for $K=0$.

\vspace{-0.075in}
\paragraph{What is not proved.} The interval verification of Theorem~\ref{thm:main}(i) relies on the accuracy of the mathematical library and on the one percent reduction of the level, so we call the inner layer verified, not proved. The hypothesis of part (iii) is tested on samples only, so the outer layer is not proved either. The direct test classifies each sample by a finite simulation, so its bound holds up to the integration error (Appendix~\ref{app:extra}). It changes no estimate, so the estimate itself is still obtained without simulating trajectories until they converge.

\vspace{-0.1in}
\section{Simulations}\label{sec:exp}
\begin{figure}[t]
\centering
\includegraphics[width=0.92\linewidth]{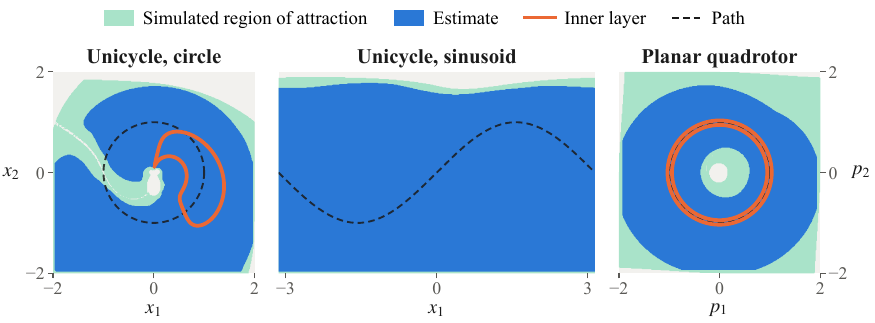}

\vspace{-0.24in}
\caption{Slices of the three cases, the unicycle at heading north and speed $1$ and the quadrotor at the orbit velocity, level tilt and hover thrust. The estimate is the union over four seeds.}
\label{fig:slices}

\vspace{-0.2in}
\end{figure}

\paragraph{Setup.} Tables~\ref{tab:config} and~\ref{tab:config2} in Appendix~\ref{app:config} list the configuration. It is the same for all systems, except for two loss weights and the shaping cost on the quadrotor (Appendix~\ref{app:config}). Four seeds are trained per case. The reference simulation classifies $5\times10^4$ quasi-random states of the box by integrating the closed loop until the state converges to $M$ or leaves $\Oad$ (Appendix~\ref{app:config}). The simulated regions of attraction hold 84.8\,\%, 95.0\,\% and 56.2\,\% of the three boxes (Figure~\ref{fig:slices}). The coverage of an estimate is the share of the reference states in the simulated region of attraction that it contains. Training, the step setting and level selection never use the reference simulation. However, we developed the configuration, including the two step settings, while watching coverage on seed 0 (Appendix~\ref{app:config}), and all reported runs, seed 0 included, were trained after these choices. The step setting keeps the long step on all three cases.

\vspace{-0.075in}
\paragraph{Baselines.} All baselines pass through level selection and the final test. The quadratic baseline is $1-e^{-\gamma\qex}$ alone. The Zubov PINN of \citet{liu2025} and the Dirichlet-Zubov PINN of \citet{mengliu2026} learn from simulated labels over four seeds. The scale of their $\tanh$ transform is searched on seed 0, and neither receives fine-tuning (Appendix~\ref{app:bench}). The no-decomposition baseline uses the same network, data, losses, level selection and fine-tuning without the term $\gamma\qex$ in the value function. The safety-value baseline sets $\gamma=0$, so it has no running cost and learns $\sup_t h$ along trajectories alone, on seed 0 without fine-tuning.

\begin{table}[t]
\centering\footnotesize\setlength{\tabcolsep}{4pt}
\caption{Coverage (\%) of the simulated region of attraction on the TFL cases, mean $\pm$ std over four seeds. Exact $v$: exact value function cut at the level of the candidate. FP: largest share of reference states inside an estimate and outside that region.}
\label{tab:main}
\begin{tabular}{@{}lccccccccc@{}}
\toprule
Case & \shortstack{Share\\of box} & Inner & Quadratic & \shortstack{Zubov\\PINN} & \shortstack{Dirichlet-\\Zubov PINN} & \shortstack{No\\decomp.} & Ours & Exact $v$ & FP\\
\midrule
Circle & 84.8 & 6.4 & 8.3 & $6.4\pm0.0$ & $11.3\pm3.9$ & $11.8\pm3.2$ & $72.0\pm1.6$ & 85.4 & 0.00\\
Sinusoid & 95.0 & 1.1 & 2.1 & $4.0\pm0.6$ & $1.1\pm0.0$ & $31.4\pm20.9$ & $83.4\pm0.8$ & 88.1 & 0.00\\
Quadrotor & 56.2 & 0.0 & 0.0 & $0.0\pm0.0$ & $0.0\pm0.0$ & $0.0\pm0.0$ & $19.1\pm4.0$ & 46.8 & 0.00\\
\bottomrule
\end{tabular}
\end{table}

\vspace{-0.075in}
\paragraph{How large are the estimates?}\looseness -1 The estimates cover most of the simulated region of attraction on the unicycle and a smaller part of it on the quadrotor (Table~\ref{tab:main}). Fine-tuning raises the mean single-seed coverage from 70.1\,\% to 72.0\,\% on the circle and from 82.1\,\% to 83.4\,\% on the sinusoid. No reference state lies inside an estimate and outside the simulated region of attraction. The direct test of unsafe states, with $10^6$ i.i.d.\ box samples per case, finds one unsafe sample in the union of four estimates of each case. With confidence $1-10^{-6}$ and up to the integration error, the states of each union outside $\D(M)$ therefore occupy at most $1.7\times10^{-5}$ of the box (Appendix~\ref{app:extra}). The quadrotor is the hardest of the three cases, with six states and a negligible inner layer (Appendix~\ref{app:extra}). Counterexamples hold its outer levels between 0.12 and 0.25. At these levels the exact value function covers 46.8\,\%, of which the outer layers reach 41\,\%, and the bound $V_\theta\ge\gamma\qex$ alone excludes 37.5\,\% of its simulated region of attraction. Both the fit and the level therefore limit this estimate.

\vspace{-0.075in}
\paragraph{Does the decomposition matter?} The candidate covers 60.2, 52.0 and 19.1 points more than the no-decomposition baseline, which must learn the integral of the running cost and covers 11.8\,\%, 31.4\,\% and 0.0\,\% under the same tests. The safety-value baseline has no running cost. On seed 0, before fine-tuning, it covers 11.1\,\%, 33.1\,\% and 1.3\,\%, against 72.0\,\%, 82.5\,\% and 24.9\,\% for the candidate (Table~\ref{tab:ablation} in Appendix~\ref{app:extra}).

\vspace{-0.075in}
\paragraph{How do the PINN baselines compare?} They cover much less than the candidate on all three cases (Table~\ref{tab:main}), although both use simulated labels and the candidate uses none. The lowest sampled value outside the inner layer belongs to a counterexample in every seed for the Zubov PINN on the circle and the quadrotor, and for the Dirichlet-Zubov PINN on the sinusoid and the quadrotor. In these cases the estimate reduces to the inner layer. However, level selection is not what limits them. Even at the level that the reference simulation itself would allow, no PINN network of any scale, seed, network size or training snapshot covers more than 20.2\,\%, 33.7\,\% and 2.4\,\% (Appendix~\ref{app:bench}).

\vspace{-0.075in}
\paragraph{Closed loops with an equilibrium.} On seed 0 of five benchmarks of the literature, the step setting of Algorithm~\ref{alg:method} keeps the long step on two and the short step on the others. After fine-tuning, the estimates cover between 86.5\,\% and 98.9\,\% of the simulated regions of attraction (Table~\ref{tab:step} in Appendix~\ref{app:bench}).

\vspace{-0.1in}
\section{Conclusions}\label{sec:conclusion}
\looseness -1 For a TFL controller with state constraints, the exact linear error dynamics make the constrained region of attraction the largest forward invariant subset of the admissible set. They also split its Zubov value function into a known quadratic cost-to-go and a safety value function. Learning only the latter by fitted value iteration gives a two-layer estimate, with an inner layer verified by interval branch and bound and an outer layer tested on samples. On average, this estimate covers 72.0\,\%, 83.4\,\% and 19.1\,\% of the simulated regions of attraction of a unicycle on two paths and of a planar quadrotor. The method needs the model, and we did not attempt the spatial quadrotor. Two further directions remain. First, a rigorous flow enclosure on the annulus would turn the test of the outer layer into a verification. Second, on the quadrotor the estimate reaches 41\,\% of what the exact value function covers at the same levels, so both a better fit and higher outer levels remain to be gained.

\noindent \textbf{Generative AI Use:} Generative AI tools were used to improve the syntax and grammar of some paragraphs in the manuscript. The authors reviewed and verified all AI-assisted content.

\bibliography{ref}

\appendix
\section{Related Work}\label{app:related}
\paragraph{Zubov functions with constraints.} \citet{camilli2001} extended Zubov's equation to perturbed systems, and \citet{gruene2015} added state constraints through an obstacle term. Our value function is their equation~(3.4) without the perturbation. \citet{xue2020} use the same kind of value function with a prescribed running cost built from a polynomial, whose integral along trajectories is unknown, and compute convex sum-of-squares inner approximations for polynomial systems. Lemma~\ref{lem:split} gives this integral in closed form for TFL closed loops. Converse Lyapunov-barrier theorems show that a smooth Lyapunov function exists on sets such as $\D(M)$ when the target set is compact and uniformly asymptotically stable, for perturbed continuous-time systems \citep{meng2022} and for hybrid systems \citep{meng2023}. Such a function is not known in closed form. Neural solvers of Zubov equations \citep{liu2025,liu2024lyznet,kang2023} and their state-constrained versions in continuous time \citep{mengliu2026} and discrete time \citep{serry2025,serry2026} learn the value function of an equilibrium or of a compact invariant set. Those that certify the result verify Lie-derivative conditions in continuous time or one-step conditions in discrete time, with satisfiability-modulo-theories or bound-propagation tools. \citet{xue2020bellman} characterize the interior of the maximal robust region of attraction of an equilibrium of a state-constrained, perturbed discrete-time system by a Bellman equation, a discrete-time analogue of the equation of \citet{gruene2015}. \citet{serry2026} note that initial estimates of the region of attraction for non-singleton sets remain open, and our inner layer gives one for TFL closed loops.

\paragraph{Safety value functions and reachability.} Equation~\eqref{eq:dp} is the self-consistency equation of an undiscounted safety value function \citep{yu2022}, and reach-avoid reinforcement learning uses a discounted form of this recursion with a target network \citep{hsu2021}. Our update follows the semi-Lagrangian scheme (5.1) of \citet{gruene2015} with such a network. In the same way, \citet{li2025} bootstrap the data term of a Zubov PINN through the network at the end of a short integration. DeepReach solves reachability equations by a residual loss \citep{bansal2021}, and \citet{bokanowski2023} use neural backward recursions for finite-horizon problems with a maximum running cost and obstacles. The closest precedent for our problem is \citet{oishi2006}, who synthesize, on a grid and in low dimension, parameterized feedback-linearizing controllers together with the set of states on which they satisfy state and input constraints. Our setting differs in the target set, the singular decoupling matrix, the learned function, and the known decrease that the running cost provides. In the language of viability theory, $\D(M)$ is the viability kernel of $\Oad$ under the closed loop.

\paragraph{Local estimates and sampled certificates.} \citet{manchester2011} computes inner estimates of the regions of attraction of hybrid limit cycles, starting from a quadratic form on the linearized transverse dynamics and enlarging it by sum-of-squares. \citet{faulwasser2016} use a sublevel set of a quadratic in transverse coordinates as the terminal region of a path-following predictive controller, for a transverse feedback linearization without a singular decoupling matrix. Finite-step Lyapunov conditions go back to \citet{aeyels1998} and \citet{geiselhart2014}. \citet{bobiti2018} verify a decrease condition on samples with Lipschitz margins on an annulus around a local set obtained from the linearization. Our annulus has the same shape and is tested on samples only, and our inner layer is verified by interval arithmetic. \citet{liu2025} combine a local quadratic region with a neural sublevel set, with the inclusion checked by a formal verifier. \citet{li2025} certify an annulus above an inner level, and \citet{score2026} certify an annulus statistically above an independently certified inner set. The anchored networks of \citet{liu2026anchored} give an inclusion of sublevel sets with a constant that depends on the parameters. Inclusion~(ii) of Theorem~\ref{thm:main} holds with no such constant. \citet{boffi2020} bound the generalization error of certificates learned from trajectories, and Proposition~\ref{prop:scen} is the one-dimensional case of \citet{campi2008}. \citet{kenanian2019} show by example that a bound on the measure of counterexamples to the decrease does not bound the measure of non-converging states. \citet{berkenkamp2016} and \citet{richards2018} close the gap between samples and the continuum with Lipschitz constants on a grid. In our setting the Lipschitz constant grows as the distances to the singular set shrink, which makes such a grid expensive in four to six dimensions.

\paragraph{Neural Lyapunov functions and TFL.} Neural Lyapunov functions with counterexample-guided training and exact verification \citep{chang2019,abate2021,dai2021,wu2023,yang2024} reach systems of six to twelve states. \citet{barreau2024} learn a maximal Lyapunov function as a quadratic plus a signed residual, optimized jointly on a prescribed compact domain. Our candidate shares this additive form. Its quadratic is fixed and exact on the admissible set, and its network is nonnegative. We found no prior network composed with an obstacle function by a maximum. The TFL literature \citep{nielsen2006,nielsen2008,akhtar2011,akhtar2015,akhtar2012,allawati2026} proves local exponential stability of the target set and treats regularity as an assumption. For the quadrotor, Assumption~A1 of \citet{allawati2026} assumes nonzero thrust on a neighbourhood of the path, and their Remark~2 notes that positive thrust is guaranteed only near the path following manifold.

\section{The Two Systems}\label{app:systems}
Both systems are instances of Section~\ref{sec:tfl} with $\ell=1$, a regulated rate of progress and $k=m=2$, so $D^\dagger=D^{-1}$, and in both the internal dynamics are the progress $\varpi$ along the path. Derivatives of circle-valued maps are computed with local angular lifts. The formulation also covers controllers that do not regulate the rate of progress. For example, the kinematic unicycle with constant speed of \citet{nielsen2006} has $n=3$, $m=1$ and one transverse output with $r_1=2$, so $k=1$, $n_z=2$, and the motion along the path at the fixed speed is internal dynamics.

\paragraph{Path following manifold.} Let $D_\perp:X_U\to\R^{\ell\times m}$ be the submatrix of $D$ formed by its rows for $\alpha_1,\dots,\alpha_\ell$, and let $X_\perp=\{x\in X_U:\operatorname{rank}D_\perp(x)=\ell\}$, an open set. On $X_\perp$ the differentials of the components of $\xi$ are linearly independent, by the argument of \citet[Section~5.1]{isidori1995} with $D_\perp$ of full row rank, so $\{x\in X_\perp:\xi(x)=0\}$ is an embedded submanifold of $X_\perp$ of dimension $n-r$. It is the maximal controlled invariant subset of $\tilde\Gamma\cap X_\perp$. Indeed, a state of $\tilde\Gamma\cap X_\perp$ with $\xi\ne0$ has a nonzero derivative $L_f^\iota\alpha_i$, $\iota<r_i$, that no input affects, so its position leaves $\Gamma$ immediately. Conversely, from a state with $\xi=0$, the feedback $u=-D_\perp^\dagger(L_f^{r_1}\alpha_1,\dots,L_f^{r_\ell}\alpha_\ell)$ keeps $\xi=0$ while the state remains in $X_\perp$. Hence $\Gamma^\star\cap X_\perp=\{x\in X_\perp:\xi(x)=0\}$, and a trajectory that stays in a compact subset of $X_\perp$ with $\xi\to0$ converges to $\Gamma^\star$. Since $\Delta_D>0$ implies that $D$, and hence $D_\perp$, has full row rank, the target set of Section~\ref{sec:roa} satisfies $M\subset\Gamma^\star$.

\paragraph{Unicycle.} The dynamically extended unicycle, with $n=4$, $r_1=r_0=2$ and $n_z=3$, a simpler model than the car-like robot of \citet{akhtar2015} with the same singularity at zero speed, has state $x=(x_1,x_2,x_3,x_4)$ and
\begin{equation}\label{eq:uni}
\dot x_1=(1+x_4)\cos x_3,\quad \dot x_2=(1+x_4)\sin x_3,\quad \dot x_3=u_2,\quad \dot x_4=u_1,
\end{equation}
so $1+x_4$ is the speed, $u_1$ the acceleration and $u_2$ the turn rate. Both outputs have relative degree two, $\xi=(\alpha,L_f\alpha)$, $\eta=(\varpi,L_f\varpi)$, $\Tm(x)=(\eta,\xi)$, and
\begin{equation}
D(x)=\begin{bmatrix}L_{g_1}L_f\varpi & L_{g_2}L_f\varpi\\ L_{g_1}L_f\alpha & L_{g_2}L_f\alpha\end{bmatrix},\qquad
\det D(x)=\pm(1+x_4)\det\begin{bmatrix}\nabla\varpi\\ \nabla\alpha\end{bmatrix}.
\end{equation}
The controller is \eqref{eq:tfl} with $\nu_\perp=-\xi_1-2\xi_2$ (double pole at $-1$) and $\nu_\parallel=-(\eta_2-1)$, so the rate of progress $\eta_2=L_f\varpi$ is driven to $\bar\eta=1$. The error coordinates are $z=(\xi_1,\xi_2,\eta_2-1)$ with $A=\operatorname{blkdiag}\big(\begin{smallmatrix}0&1\\-1&-2\end{smallmatrix}\big),-1\big)$ and $Q=\operatorname{diag}(1.742,0.249,3.281)$. The target set $M=\{z=0\}$ is the path traversed at the rate of progress $\eta_2=1$. For the unicycle it has two components, because $\eta_2$ depends on the velocity and not on the heading. In the first the heading is along the path and the speed is positive, and in the second the heading is against the path and the speed is negative. On both components, $|\det D|$ and $|\det J_\Tm|$ are bounded away from zero and the state is inside the box, as the expressions below show, so $M$ is compact and lies in $\Oad$. The reference simulation counts convergence to either component.

\emph{Circle.} The path is $\alpha=x_1^2+x_2^2-1$ with $\varpi=\operatorname{atan2}(x_2,x_1)$, so $\xi_2=2(1+x_4)(x_1\cos x_3+x_2\sin x_3)$, $\eta_2=(1+x_4)(x_1\sin x_3-x_2\cos x_3)/(x_1^2+x_2^2)$, $\det D=-2(1+x_4)$, $\det J_\Tm=4(1+x_4)$, $\|\nabla\varpi\|^{-1}=\sqrt{x_1^2+x_2^2}$. Box: $|x_1|,|x_2|\le2$, $x_4\in[-3,1]$ (speed in $[-2,2]$), $x_3\in S^1$. \emph{Sinusoid.} The path is $\alpha=x_2-\sin x_1$ with $\varpi=x_1$, so $\xi_2=(1+x_4)(\sin x_3-\cos x_1\cos x_3)$, $\eta_2=(1+x_4)\cos x_3$, $\det D=\det J_\Tm=1+x_4$, $\|\nabla\varpi\|^{-1}=1$. The closed loop is $2\pi$-periodic in $x_1$, taken in $[-\pi,\pi)$. The box is $|x_2|\le2$, $x_4\in[-3,1]$. On $M$ the speed has magnitude $\sqrt{1+\cos^2x_1}\le\sqrt2$. The thresholds are $\delta_D=\delta_\Tm=10^{-2}$ and $\delta_\varpi=5\cdot10^{-2}$, so $\Oad$ excludes $|1+x_4|\le0.005$ (circle) or $\le0.01$ (sinusoid) and, for the circle, the disc of radius $0.05$ around the centre.

\paragraph{Planar quadrotor.} The planar quadrotor, with $n=6$, $r_1=r_0=3$ and $n_z=5$, has state $(p_1,p_2,\dot p_1,\dot p_2,\Theta,T)$ and
\begin{equation}\label{eq:quad}
\ddot p=T\begin{bmatrix}-\sin\Theta\\ \cos\Theta\end{bmatrix}-\begin{bmatrix}0\\1\end{bmatrix},\qquad
\dot\Theta=u_2,\qquad \dot T=u_1,
\end{equation}
in units with unit mass, unit gravity and unit path radius. The thrust $T$ is a controller state (the dynamic extension of \citet{akhtar2012}), and the tilt rate is the second input, as provided by a multirotor autopilot with an inner rate loop. For the loop $\alpha=p_1^2+p_2^2-1$, $\varpi=\operatorname{atan2}(p_2,p_1)$ both outputs have relative degree three: $\xi=(\alpha,\ 2p^\top\dot p,\ 2|\dot p|^2+2p^\top\ddot p)$, $\eta_2=(p_1\dot p_2-p_2\dot p_1)/|p|^2$ the angular rate, $\eta_3=L_f^2\varpi$, and
\begin{equation}
D(x)=\begin{bmatrix}\upsilon_1/|p|^2 & T\upsilon_2/|p|^2\\ 2\upsilon_2 & -2T\upsilon_1\end{bmatrix},\qquad
\upsilon_1=p_1\cos\Theta+p_2\sin\Theta,\quad \upsilon_2=p_2\cos\Theta-p_1\sin\Theta,
\end{equation}
so $\det D=-2T$ and $\det J_\Tm=-8T$. The controller is \eqref{eq:tfl} with $\nu_\perp=-8\xi_1-12\xi_2-6\xi_3$ (triple pole at $-2$) and $\nu_\parallel=-(\eta_2-\tfrac12)-2\eta_3$ (double pole at $-1$). The error coordinates are $z=(\xi_1,\xi_2,\xi_3,\eta_2-\tfrac12,\eta_3)$ with $Q=I$, and $M=\{z=0\}$ is the loop traversed counter-clockwise at angular rate $\tfrac12$. On $M$ the thrust is $T=|\ddot p+(0,1)|\in[0.75,1.25]$. The box is $|p_i|\le2$, $|\dot p_i|\le2$, $T\in[0,4]$ (thrust-to-weight ratio up to four), $\Theta\in S^1$. The thresholds are those of the unicycle, so $\Oad$ excludes $T\le0.005$ and the disc of radius $0.05$ around the centre. The reference simulation treats a sign change of $T$ between two steps as a departure through the singular set.

\paragraph{Constants.} Table~\ref{tab:consts} lists the scaling $S_z$ and the cost scale $\gamma$. The scale is $\gamma=1/q_{90}$ for the long-step setting and $\gamma=3/q_{90}$ for the short-step setting of Section~\ref{sec:train}, where $q_{90}$ is the 90th percentile of $\qex$ over admissible samples. Both constants are computed from the model, without simulating trajectories.
\begin{table}[ht]
\centering\small
\caption{Constants computed before training: the cost scale $\gamma$ of the long-step setting and the diagonal of $S_z^{-1}$.}
\label{tab:consts}
\begin{tabular}{@{}lcl@{}}
\toprule
Case & $\gamma$ & $S_z^{-1}$\\
\midrule
Circle & 0.0444 & 1.550, 1.396, 1.011\\
Sinusoid & 0.0608 & 1.001, 0.625, 1.000\\
Quadrotor & 0.0288 & 1.546, 2.439, 5.953, 0.703, 1.034\\
\bottomrule
\end{tabular}
\end{table}

\section{Proofs}\label{app:proofs}
\paragraph{The target set.} Since $\overline{\Oad}$ is compact and $z$ is continuous, $M$ is compact. A trajectory from $M$ keeps $z=0$ while it stays in $\Oad$. It cannot reach $\partial\Oad$, because the first such point would lie in $M\subset\Oad$ by continuity of $z$ and Assumption~\ref{ass:main}(c). Hence $M$ is forward invariant.

\paragraph{Proof of Lemma~\ref{lem:split}.} We prove the lemma with the shaping cost of Appendix~\ref{app:config}, for which $C_t(x)=\kappa\int_0^t\rho(y(s;x))\,ds$, and $\kappa=0$ gives the statement. By Assumption~\ref{ass:main}(b), $\dot z=Az$ on $\Oad$, hence $\dot\zeta=S_z\dot z=\tilde A\zeta$ and $L_{f_{\mathrm{cl}}}\qex=\zeta^\top(\tilde A^\top P+P\tilde A)\zeta=-\zeta^\top Q\zeta$. Integrating along a trajectory that stays in $\Oad$ on $[0,t]$ gives $\int_0^t\mathcal L(y(s;x))\,ds=\gamma\qex(x)-\gamma\qex(y(t;x))$. Let the trajectory of $x$ stay in $\Oad$ for all $t\ge0$. Substituting in \eqref{eq:V} with the shaping cost, $V(x)=\gamma\qex(x)+\sup_{t\ge0}g(t)$ with $g(t)=C_t(x)+(h-\gamma\qex)(y(t;x))$. By Theorem~\ref{thm:gz}, $\dist(y(t;x),M)\to0$ and $z(y(t;x))\to0$. Since $h$ vanishes on a neighbourhood of the compact set $M$ and $\rho$ vanishes on $\{\gamma\qex<c_I/2\}$, $h(y(t;x))=0$ and $\rho(y(t;x))=0$ for all large $t$, so $C_t(x)$ converges to a finite $C_\infty(x)$ and $g(t)\to C_\infty(x)$. Replacing $h-\gamma\qex$ by $\psi=\relu(h-\gamma\qex)$ changes $g(t)$ only where $h-\gamma\qex<0$, and there it gives $C_t(x)\le C_\infty(x)\le\sup_sg(s)$. Hence $\sup_tg(t)=\sup_t[C_t(x)+\psi(y(t;x))]=\Phi(x)$, which is \eqref{eq:split} for $\kappa=0$. If the trajectory leaves $\Oad$, both sides are infinite by definition. Since $C_t\ge0$ and $\psi\ge0$, $\Phi\ge\psi\ge0$ and $V\ge\gamma\qex$. For \eqref{eq:dp} and \eqref{eq:dpk}, split the supremum over $[0,\tau]$ and $[\tau,\infty)$ and use $y(t;x)=y(t-\tau;y(\tau;x))$ and $C_t(x)=C_\tau(x)+C_{t-\tau}(y(\tau;x))$ for $t\ge\tau$. In normalized time $s$, $dt=ds/a$, so the same identities hold along the orbit of $\bar f$ with $\bar C_s$ in place of $C_t$. The convergence part of the proof of Theorem~\ref{thm:gz} uses only Assumption~\ref{ass:main}, and its finiteness part uses only the first identity of this lemma, so the argument is not circular. \hfill$\square$

Substituting \eqref{eq:split} in the obstacle equation of Theorem~4.4 of \citet{gruene2015} gives, formally, the variational inequality $\min\{-L_{f_{\mathrm{cl}}}\Phi-\kappa\rho,\ \Phi-\psi\}=0$ on $\Oad$. Identity \eqref{eq:dp} is the dynamic programming principle (4.11) of \citet{gruene2015} written for $\Phi$. While $\Phi$ is only nonincreasing along trajectories in $\Oad$, $V$ keeps the strict decrease of Remark~4.3 of \citet{gruene2015}: for every $x\in\D(M)$ and $t\ge0$, $V(y(t;x))\le V(x)-\gamma\big(\qex(x)-\qex(y(t;x))\big)$, hence $V(y(t;x))<V(x)$ when $x\notin M$ and $t>0$.

\paragraph{Proof of Theorem~\ref{thm:gz}.} Let the trajectory of $x\in\Oad$ stay in $\Oad$ while it exists. By Assumption~\ref{ass:main}(a), $f_{\mathrm{cl}}$ is bounded on the compact set $\overline{\Oad}$, so the solution exists for all $t\ge0$. By Assumption~\ref{ass:main}(b), $z(y(t;x))=e^{At}z(x)\to0$. If $\dist(y(t_j;x),M)$ stays bounded away from zero along a sequence $t_j\to\infty$, a subsequence of $y(t_j;x)$ converges to a point of $\overline{\Oad}$ with $z=0$, that is, a point of $M$. This point would lie at positive distance from $M$, which is impossible. Hence $\dist(y(t;x),M)\to0$ and $x\in\D(M)$. This argument uses no property of the internal dynamics. The converse holds by the definition of $\D(M)$. A forward invariant subset of $\Oad$ consists of states whose trajectories never leave $\Oad$, which gives the maximality. Let $x\in\D(M)$. The closure of its trajectory consists of the trajectory and its limit points, which lie in $M\subset\Oad$ by Assumption~\ref{ass:main}(c). It is therefore a compact subset of $\Oad$, and the continuous function $h$ is bounded on it. By the first identity of Lemma~\ref{lem:split}, $\int_0^t\mathcal L(y(s;x))\,ds\le\gamma\qex(x)$ for all $t$. The shaping cost vanishes on $\{\gamma\qex<c_I/2\}$, which the trajectory enters and does not leave because $\qex(y(t;x))$ is nonincreasing and tends to zero, so its integral is finite. Hence $V(x)<\infty$. Conversely, $V(x)<\infty$ implies by definition that the trajectory of $x$ stays in $\Oad$. \hfill$\square$

\paragraph{Proof of Theorem~\ref{thm:main}.} (i) Let $x\in\I$. By hypothesis, $x$ satisfies the regularity constraints, so $x\in\Oad$. Let $t_*=\sup\{t\ge0:y(s;x)\in\Oad\text{ for all }s\in[0,t]\}$. By Lemma~\ref{lem:split}, $\qex$ is nonincreasing along the trajectory on $[0,t_*)$, so $\gamma\qex(y(t;x))\le\gamma\qex(x)<c_I$ there. Suppose $t_*<\infty$. Since $\overline{\Oad}$ is compact and $f_{\mathrm{cl}}$ is smooth on a neighbourhood of it, $y(t_*;x)$ exists, lies in $\partial\Oad\subset B\cap X_U$, and has $\gamma\qex<c_I$ by continuity. By hypothesis it satisfies the regularity constraints, so it lies on a nonperiodic face of $B$, and $f_{\mathrm{cl}}$ points strictly into $B$ across every active face there. Then the trajectory lies outside $B$ just before $t_*$, which contradicts $y(t;x)\in B^\circ$ for $t<t_*$. Hence $t_*=\infty$, the trajectory stays in $\Oad\cap\{\gamma\qex<c_I\}\subset\I$, so $\I$ is forward invariant, and $\I\subset\D(M)$ by Theorem~\ref{thm:gz}. The points of $M$ have $\qex=0$ and lie in $\Oad\subset B^\circ\cap X_U$, so $M\subset\I$. (ii) The inequality $\tilde v<c_1$ means $-\ln(1-\tilde v)<c_I$, and $\gamma\qex\le-\ln(1-\tilde v)$. (iii) Let $x\in\Oad$ with $\tilde v(x)<c_2$, and define states $x^{(0)}=x,x^{(1)},\dots$ as follows. If $x^{(i)}\in\I$ or $\tilde v(x^{(i)})<c_1$, which implies $x^{(i)}\in\I$ by (ii), stop. Otherwise $x^{(i)}\in\Oad\setminus\I$ and $c_1\le\tilde v(x^{(i)})<c_2$, so for some $H_i\in\{1,2,4,8\}$ the orbit of $\bar f$ from $x^{(i)}$ stays in $\Oad$ over $[0,H_i]$ and $x^{(i+1)}=\bar y(H_i;x^{(i)})$ has $\tilde v(x^{(i+1)})<\tilde v(x^{(i)})<c_2$. If the construction stops at some $x^{(i)}\in\I$, the orbit stays in $\Oad$ afterwards because $\I$ is forward invariant by (i). Otherwise the steps, each of length at least one, cover all normalized time. In both cases the orbit of $\bar f$ from $x$ stays in $\Oad$ for all normalized time. By Assumption~\ref{ass:main}(a), $a$ is bounded on $\overline{\Oad}$ by some $a_{\max}$, so the physical time $t(s)=\int_0^s a(\bar y(s';x))^{-1}\,ds'\ge s/a_{\max}$ diverges, the trajectory of $f_{\mathrm{cl}}$ stays in $\Oad$ for all $t\ge0$, and $x\in\D(M)$ by Theorem~\ref{thm:gz}. \hfill$\square$

\begin{proposition}\label{prop:scen}
Fix a measurable function $\tilde v:B\to[0,1]$, levels $0<c_1\le\bar c_2\le1$ and a measurable counterexample test before drawing $N$ i.i.d.\ samples from the uniform distribution $\mathbb P$ on $B$. Let $c_2$ be the minimum of $\bar c_2$ and the values of $\tilde v$ at the sampled counterexamples with $\tilde v\ge c_1$, and let $\mathcal C$ be the set of counterexamples of the test, as implemented, in $\{c_1\le\tilde v<c_2\}$. Then $\mathbb P^N\big[\mathbb P(\mathcal C)>\varepsilon\big]\le(1-\varepsilon)^N$ for every $\varepsilon\in(0,1)$.
\end{proposition}

\paragraph{Proof of Proposition~\ref{prop:scen}.} Let $G(c)=\mathbb P(\text{counterexamples in }\{c_1\le\tilde v<c\})$, nondecreasing and left-continuous. If $G(\bar c_2)\le\varepsilon$ there is nothing to prove. Otherwise let $c_*=\inf\{c:G(c)>\varepsilon\}$. The set $\mathcal C_*$ of counterexamples in $\{c_1\le\tilde v\le c_*\}$ has measure at least $\varepsilon$, the event $G(c_2)>\varepsilon$ requires $c_2>c_*$ and hence no sample in $\mathcal C_*$, and that has probability at most $(1-\varepsilon)^N$. This is the one-dimensional case of the scenario theorem of \citet{campi2008}. \hfill$\square$

\begin{proposition}\label{prop:unsafe}
Let the hypotheses of Theorem~\ref{thm:main}(i) hold, and fix a measurable set $\mathcal E\subset B$ and a horizon $t_{\max}>0$ in normalized time before drawing $N$ i.i.d.\ samples from $\mathbb P$. Let
\[
\mathcal E_{\mathrm u}=\mathcal E\setminus\big\{x\in\Oad:\ \exists\,t\in[0,t_{\max}]\ \text{with}\ \bar y(t;x)\in\I\ \text{and}\ \bar y(s;x)\in\Oad\ \text{for all}\ s\in[0,t]\big\}.
\]
Then $\mathcal E\setminus\D(M)\subset\mathcal E_{\mathrm u}$. Let $K$ be the number of samples in $\mathcal E_{\mathrm u}$, and for $j<N$ let $\varepsilon_j$ solve $\Pr[\mathrm{Bin}(N,\varepsilon_j)\le j]=10^{-6}$, with $\varepsilon_N=1$. Then $\mathbb P^N\big[\mathbb P(\mathcal E_{\mathrm u})>\varepsilon_K\big]\le10^{-6}$.
\end{proposition}

\paragraph{Proof of Proposition~\ref{prop:unsafe}.} A state of $\mathcal E\setminus\mathcal E_{\mathrm u}$ lies in $\Oad$ and reaches $\I$ along a trajectory in $\Oad$. By Theorem~\ref{thm:main}(i), $\I$ is forward invariant and lies in $\D(M)$, so the state lies in $\D(M)$, and $\mathcal E\setminus\D(M)\subset\mathcal E_{\mathrm u}$. The count $K$ of samples in $\mathcal E_{\mathrm u}$ is binomial with parameters $N$ and $\bar\varepsilon=\mathbb P(\mathcal E_{\mathrm u})$. Since $\varepsilon_j$ increases with $j$, the event $\bar\varepsilon>\varepsilon_K$ is the event $K\le j_*$, with $j_*$ the largest $j$ such that $\varepsilon_j<\bar\varepsilon$, and it is empty if there is no such $j$. The distribution function of the binomial at $j_*$ decreases in its success probability, so $\Pr[K\le j_*]\le\Pr[\mathrm{Bin}(N,\varepsilon_{j_*})\le j_*]=10^{-6}$. This is the one-sided Clopper-Pearson bound. \hfill$\square$

\section{Baselines and Closed Loops with an Equilibrium}\label{app:bench}
\paragraph{Zubov PINN.} The Zubov PINN follows \citet{liu2025}: two hidden layers of 30 tanh units, $W=\tanh(\sigma V)$, residual $\nabla W\cdot f+\sigma\omega(1-W^2)=0$, a boundary term, and a data term on $3000$ simulated states. It is trained for $2\times10^4$ Adam steps on batches of $2048$ from $10^5$ collocation points. On the TFL closed loops the residual is divided by $a(x)$ of Section~\ref{sec:problem}, and the running cost is $\omega=\zeta^\top Q\zeta$. The inputs are the normalized state with angles as sine and cosine, and four snapshots are screened. Both PINNs are set to one outside $\Oad$.

\paragraph{Dirichlet-Zubov PINN.} We follow \citet{mengliu2026}. With $d_{\min}=\min_id_i/\mathrm{median}\,d_i$ and their scaling constant $0.1$, the same residual is solved with the field $0.1\,d_{\min}f$. The function is zero at the target and one on the $2000$ admissible states closest to the boundary of $\Oad$. The data term integrates $\omega/(0.1\,d_{\min})$ along the simulated trajectories, collocation and the boundary term act on admissible states, and the residual is divided by $\sqrt{1+\|0.1\,d_{\min}Sf\|^2+(\sigma\omega)^2}$, a positive weight.

\paragraph{Scale search.} Both PINNs use $\sigma=3/(\varsigma\,q_{90})$, with $q_{90}$ as in the definition of $\gamma$. The divisor $\varsigma$ is searched on seed 0 by the coverage after level selection and the final test, with the scoring of the reported numbers. For the Zubov PINN the grid is $\{1,3,10,30,100,300\}$. For the Dirichlet-Zubov PINN it is $\{1,3,10,30,100\}$, extended to $300$ when $100$ is best and to $1000$ when $300$ is best. Ties go to the smallest $\varsigma$. On the circle and the quadrotor the Zubov PINN ties at the inner layer at every scale, as does the Dirichlet-Zubov PINN on the sinusoid and the quadrotor. On the sinusoid, the smaller divisor $\varsigma=0.3$ gives no gain for the Zubov PINN.

\paragraph{Larger PINNs and the oracle level.} On the TFL closed loops we also trained both PINNs on seed 0 with the features and the network of the candidate, at $\varsigma=1$ and $\varsigma=10$. The best of these four runs covers 18.2\,\%, 2.5\,\% and 0.0\,\% on the three cases. To bound what any choice of level could give the PINNs there, we also compute an oracle level for every training snapshot of every PINN run. These runs are those of the scale searches, the four seeds and the larger networks. It is the largest level at which no reference state outside the simulated region of attraction lies in the estimate, so it uses the reference labels that the method never sees. The best of these estimates covers 20.2\,\%, 33.7\,\% and 2.4\,\% on the three cases.

\paragraph{Other baselines on the TFL closed loops.} The no-decomposition baseline keeps the features of the candidate and its factor $\tanh(\qex/q_0)$. Its function is $V_\theta=\max\{h,\Psi_\theta\}$, and the running cost $\mathcal L$ is integrated along the steps and included in the target. Like sum-of-squares \citep{xue2020}, moment methods \citep{korda2013} require polynomial dynamics and are not applied to the TFL closed loops (Section~\ref{sec:intro}).

\paragraph{Scoring.} We score our reimplementations of both PINNs with the level selection and the final test of this paper. For the PINNs, the no-decomposition baseline and the safety-value baseline, $c_1$ is set on samples, because their functions have no lower bound $\gamma\qex$. The PINNs are not clipped at zero, because clipping creates plateaus on which no step can decrease, and the oracle level above bounds what any choice of level could give them.

\paragraph{Closed loops with an equilibrium.} Algorithm~\ref{alg:method} also applies to a closed loop $\dot x=f(x)$ with an exponentially stable equilibrium at the origin. Its admissible set $\Oad$ consists of the states of $B^\circ$ that satisfy the state constraints and avoid the obstacles, and $h$ is built as in \eqref{eq:h}. We take $q(x)=\tilde x^\top P\tilde x$ with $\tilde x=Sx$, where $P$ solves $\tilde A^\top P+P\tilde A=-I$ for $\tilde A=S\,\tfrac{\partial f}{\partial x}(0)\,S^{-1}$, and the running cost $\mathcal L=\gamma\|\tilde x\|^2+\kappa\rho$, where $\gamma$ is that of the step setting, with $q_{90}$ the 90th percentile of $q$ over admissible samples. Since $L_fq\neq-\|\tilde x\|^2$ away from the origin, the integral of $\mathcal L$ is no longer known in closed form. The candidate \eqref{eq:ansatz} and the target with an accumulated cost (Appendix~\ref{app:config}) therefore use $q$ in place of $\qex$ and $C_s=\int_0^s\mathcal L\,dt+\gamma(q(y(s))-q(x))$, and $\tilde x$ replaces $\zeta$ in the features. For the same reason, $C_s$ can decrease along a trajectory, so the fixed point of the target can overestimate $\Phi$, for every step (Appendix~\ref{app:config}). The network, the training and the fine-tuning are otherwise unchanged. On these systems the inner layer is a sublevel set of $q$ at a level set on samples, so it is not verified. Moreover, Theorem~\ref{thm:gz} is not available, because a trajectory that stays in $\Oad$ need not converge. The estimates are therefore tested on samples only.

\paragraph{Systems.} Each system has its equilibrium at the origin. Leaving the box counts as leaving $\Oad$. The reversed Van der Pol oscillator is $\dot x_1=-x_2$, $\dot x_2=x_1-\mu(1-x_1^2)x_2$, on $[-2.5,2.5]\times[-3.5,3.5]$ for $\mu=1$ and $[-3,3]\times[-6,6]$ for $\mu=3$. The pendulum is $\dot x_1=x_2$, $\dot x_2=\sin x_1-x_2+u$, with $u=\operatorname{sat}_{0.8}(-4.4142x_1-2.3163x_2)$, the gains of Example~5 of \citet{liu2025} and a torque bound that we add, on $[-\pi,\pi]\times[-4,4]$. The 3-D system is $\dot x_1=-x_1+x_2$, $\dot x_2=0.1x_1-2x_2-x_1^2-0.1x_1^3$, $\dot x_3=-x_3+x_1x_2$, with $x_3<1$ and $(x_1+1)^2+(x_2-1)^2>0.3^2$, on $[-3,3]^2\times[-2,2]$. The power system is $\dot x_1=x_2$, $\dot x_2=-0.5x_2-\sin(x_1+\pi/3)+\sin(\pi/3)$, on $[-2,3]\times[-3,3]$. The reference simulation integrates $5\times10^4$ states with RK4, step $0.01$ and horizon $40$ ($60$ for the power system), and a state has converged when the scaled state has norm below $10^{-3}$ for one time unit.

\paragraph{Results.} We ran Algorithm~\ref{alg:method} on seed 0 of the five systems (Table~\ref{tab:step}). Its step setting keeps the long step on two of them and the short step on the others. On every system, as on the TFL cases, the share of box samples ranks the two settings as the coverage does. After fine-tuning, the estimates of the kept setting cover between 86.5\,\% and 98.9\,\% of the simulated regions of attraction, and no reference state lies inside one of them and outside the simulated region of attraction. We ran one seed per system and did not compare with baselines on these systems.

\begin{table}[ht]
\centering\small
\caption{Step setting of Algorithm~\ref{alg:method} on seed 0. Share: share of box samples in the estimate (\%), before fine-tuning, which decides the setting without reference labels. Coverage (\%) of the simulated region of attraction is shown for comparison, before fine-tuning for both settings and after fine-tuning for the kept one.}
\label{tab:step}
\begin{tabular}{@{}lcccccc@{}}
\toprule
System & \shortstack{Share,\\long} & \shortstack{Share,\\short} & Kept & \shortstack{Coverage,\\long} & \shortstack{Coverage,\\short} & \shortstack{Coverage,\\kept, fine-tuned}\\
\midrule
Circle & 61.1 & 59.3 & long & 72.0 & 69.8 & 74.0\\
Sinusoid & 78.4 & 54.3 & long & 82.5 & 57.2 & 83.7\\
Quadrotor & 14.0 & 5.3 & long & 24.9 & 9.4 & 22.9\\
\midrule
Van der Pol, $\mu=1$ & 30.3 & 38.8 & short & 77.3 & 98.9 & 98.9\\
Van der Pol, $\mu=3$ & 15.4 & 21.3 & short & 54.1 & 74.9 & 89.2\\
Pendulum & 30.2 & 27.4 & long & 93.5 & 84.6 & 94.2\\
3-D system & 50.0 & 40.0 & long & 84.7 & 67.8 & 86.5\\
Power system & 15.7 & 29.9 & short & 47.5 & 90.3 & 92.4\\
\bottomrule
\end{tabular}
\end{table}

\section{Interval Verification of the Inner Layer}\label{app:inner}
The hypotheses of Theorem~\ref{thm:main}(i) are statements about explicit functions on a box, and we check them by interval branch and bound \citep{moore2009}. The box is split into cells. A cell is discarded when interval arithmetic shows that $\gamma\qex\ge c_I$ on it or that it contains no violating point. A cell contains no violating point when all distances $d_i$ to the regularity constraints are positive on it and, if it touches a face, the normal component of the field has the inward sign. The remaining cells are bisected along their widest normalized side until none is left. The checks run on the closed box with the periodic angle over $[-\pi,\pi]$. The lower bound of the quadratic form is the larger of two bounds: the one from the eigendecomposition of $P$, with each squared projection of $\zeta$ bounded below by zero, and $\lambda_{\min}(P)\sum_i\zeta_i^2$. The distances $d_i$ and the field components are the symbolic expressions of Appendix~\ref{app:systems}, evaluated in float64 interval arithmetic with outward rounding by two units in the last place per operation and with enclosed constants. For each case the level is found by bisection over twelve rounds between zero and the sampled minimum of $\gamma\qex$ over samples outside $\Oad$ and samples on box faces where the field points outward. The last verified level is multiplied by $0.99$. This one percent reduction absorbs the float32 evaluation of $\qex$ in the training pipeline and the residual effects of the eigendecomposition of $P$ and the float32 storage of $P$ and $S_z$. Table~\ref{tab:inner} lists the verified levels, the share of the box in $\I$, the number of cells and the time. The verified claim concerns the set $\{\qex<c_I/\gamma\}$ and does not otherwise involve $\gamma$. The verification was run with the cost scale $3/q_{90}$ of the short-step setting, and for the long-step setting the level $c_I$ is one third of the verified one, which describes the same set. Table~\ref{tab:inner} gives the levels for $\gamma=1/q_{90}$.
\begin{table}[ht]
\centering\small
\caption{Verified inner layers of the TFL cases, with the cells and time of the final verification.}
\label{tab:inner}
\begin{tabular}{lccccc}
\toprule
Case & $c_I$ & $c_1$ & $\I$ (\% box) & Cells & Time (s)\\
\midrule
Circle & 0.0350 & 0.0344 & 5.40 & 944395 & 588\\
Sinusoid & 0.0170 & 0.0168 & 1.01 & 1047557 & 110\\
Quadrotor & 0.0002 & 0.0002 & 0.00 & 21270981 & 3626\\
\bottomrule
\end{tabular}
\end{table}

\section{Configuration}\label{app:config}
\paragraph{Training setup.} The candidate uses the network, optimizer and collocation points of Table~\ref{tab:config}, with the level selection and fine-tuning of Section~\ref{sec:method}. We set the configuration on seed 0, with level selection and a final test, while watching the coverage of the reference simulation. In a first stage we chose the weight of the finite-step penalty and $V_{\max}$, which gave the short-step setting. In a second stage, with these weights fixed, we chose the step $\tau$, the cost scale $\gamma$ and the exit weight $w_{\mathrm{ex}}$ one at a time on the quadrotor, which gave the long-step setting. The other constants of Tables~\ref{tab:config} and~\ref{tab:config2} come from an earlier version of the method on the unicycle. All reported runs, seed 0 included, were trained after these choices. Since the reference simulation guided these choices on seed 0, the coverage of seed 0 is not an independent test of the configuration, and Table~\ref{tab:seeds} lists every seed.

\begin{table}[t]
\centering\footnotesize\renewcommand{\arraystretch}{1.1}
\caption{Configuration of every case: value function, candidate and training. Step lengths are in normalized time.}
\label{tab:config}
\begin{tabular}{@{}>{\raggedright\arraybackslash}p{0.16\linewidth}>{\raggedright\arraybackslash}p{0.81\linewidth}@{}}
\toprule
\multicolumn{2}{@{}l}{\textit{Value function and candidate}}\\
Value function & $b_i=0.1\cdot$median $d_i$, $h=\sum_i\relu(b_i/d_i-1)^2$, $\gamma$ of the step setting, with $q_{90}$ the 90th percentile of $\qex$ over admissible samples\\
Shaping cost & $\kappa=0.3$ if the inner layer holds at least $1\,\%$ of $2\cdot10^5$ box samples, and $\kappa=0$ otherwise. The function $\rho$ is a smooth step from $\gamma\qex=c_I/2$ to $c_I$. The share on the sinusoid is $1.01\,\%$, close to this threshold\\
Network & $\mathcal N_\theta$: 64-64-64 tanh. Features: normalized state (angles as sine and cosine), $\tanh(\zeta/3)$, $\tanh(\tfrac12\ln(d_i/\mathrm{median}\,d_i))$ for the constraints on $\Delta_D$ and $\|\nabla\varpi\|^{-1}$. Softplus sharpness $\beta=10$. The constant $q_0$ is the first percentile of $\qex$ on $2\cdot10^5$ collocation points\\
Initialization & The network starts close to $\Phi_\theta=\psi$, which lies below the exact solution\\
\midrule
\multicolumn{2}{@{}l}{\textit{Training}}\\
Collocation & $4\cdot10^5$ Sobol box samples, $10^5$ near the faces, $10^5$ next to the singular set (admissible ones kept)\\
Step setting & Long: $\tau=2$ (80 RK4 substeps), $\gamma=1/q_{90}$, $w_{\mathrm{ex}}=10$. Short: $\tau=0.2$ (8 RK4 substeps), $\gamma=3/q_{90}$, $w_{\mathrm{ex}}=0$. Each system keeps the setting whose seed-0 estimate, after level selection and the final test and before fine-tuning, holds the larger share of box samples. The regression loss and the obstacle penalty of a point whose step leaves $\Oad$ are multiplied by $1+w_{\mathrm{ex}}$\\
Target network & Copy of the averaged weights (decay $0.999$), refreshed every epoch\\
Optimization & Adam, 100 epochs, batch 4096, cosine learning rate $10^{-3}\to10^{-5}$. Exponential loss weight $e^{2\min(V_\theta,V_{\max})}$ with $V_{\max}=3$ (no such weight on the quadrotor)\\
Decrease penalty & Weight 10, on $\relu((L_{f_{\mathrm{cl}}}\Phi_\theta+\kappa\rho)/a)\,e^{-V_\theta}$, where $\Psi_\theta>\psi$\\
Obstacle penalty & Weight 10, on $\relu(\psi+0.01-\Psi_\theta)\,e^{-V_\theta}$, where $\Psi_\theta\le\psi$ and $L_{f_{\mathrm{cl}}}\psi+\kappa\rho>0$\\
Finite-step penalty & Weight 100 (10 on the quadrotor), unit step with 40 RK4 substeps, on $\relu(e^{-(V_\theta(x)-\chi[\gamma\relu(\qex(x)-\qex(x^{+}))+C_1])}-e^{-V_\theta(x^{+})})$ with $\chi=0.2$ and $x^{+}$ the end of the unit step from $x$, on steps in $\Oad$ with $V_\theta(x)<8$\\
Violating states & Every 5 epochs up to epoch 90: the $10^4$ states whose unit step violates the decrease most (no margin), in a pool of $2\cdot10^5$ box samples and $5\cdot10^4$ next to the singular set, 4 Gaussian neighbours (standard deviation $0.01$ of the half-widths), rolling buffer $10^5$\\
\bottomrule
\end{tabular}
\end{table}
\begin{table}[t]
\centering\footnotesize\renewcommand{\arraystretch}{1.1}
\caption{Configuration of every case: level selection, fine-tuning and reference simulation. Step lengths are in normalized time, except in the reference simulation.}
\label{tab:config2}
\begin{tabular}{@{}>{\raggedright\arraybackslash}p{0.16\linewidth}>{\raggedright\arraybackslash}p{0.81\linewidth}@{}}
\toprule
\multicolumn{2}{@{}l}{\textit{Level selection and fine-tuning}}\\
Level selection & Horizons $H\in\{1,2,4,8\}$, RK4 substep $0.025$, states with $\|\zeta\|\le10^{-3}$, which lie in $\I$, exempt from the decrease. Snapshots every 5 epochs and at epoch 0. A test on a number of box samples adds half as many near the faces, a fifth as many next to the singular set and half as many on the faces, where states with outward field count as counterexamples. Screening: $5\cdot10^4$ box samples, 3 rounds. Retest of the top 3: $10^5$ box samples, 5 rounds. Ranking by share of $5\cdot10^4$ fixed box samples. Final test: $10^6$ i.i.d.\ uniform box samples\\
Fine-tuning & 4 rounds, each on the counterexamples with $c_1\le v_\theta<\min\{c_2+0.05,\,0.999\}$ for a level $c_2$ estimated again in that round, pool of $10^6$ box samples, $5\cdot10^5$ near the faces and $2\cdot10^5$ next to the singular set, 8 Gaussian neighbours per counterexample, 300 Adam steps (learning rate $5\cdot10^{-5}$, batch 4096), weight 100 on counterexamples whose step stays in $\Oad$, weight 10 on those that leave it, decrease margin 0.2, counterexample sets capped at $5\cdot10^4$, anchor on $2\cdot10^5$ box samples, exponential loss weight with $V_{\max}=3$ on all systems\\
\midrule
\multicolumn{2}{@{}l}{\textit{Reference simulation}}\\
Reference & $5\cdot10^4$ Sobol states, RK4 step 0.01, horizon 30 (unicycle) and 40 (quadrotor). Converged: held for one time unit, $\|\xi\|<10^{-3}$ and $|\eta_2-1|<0.05$ (unicycle), $\|\zeta\|<10^{-3}$ (quadrotor). States that have neither converged nor failed at the horizon count as outside\\
\bottomrule
\end{tabular}
\end{table}

\paragraph{Step setting.} Unrolling the target over successive steps, with the shaping cost below, shows that its only continuous fixed point that vanishes on $M$ is $\sup_{t\ge0}[C_t(x)+\psi(y(t;x))]$ for every step $\tau$. The candidate satisfies both conditions, because $\Psi_\theta$ and $\psi$ vanish on $M$. On a TFL closed loop this function is the safety value function $\Phi=V-\gamma\qex$. Indeed, $\qex$ is the exact cost-to-go of the error dynamics, so the cost accumulated beyond $\gamma\qex$ is the nonnegative shaping cost alone. Hence $C_t$ is nondecreasing in $t$, and the margin function $\psi=\relu(h-\gamma\qex)$ can replace $h-\gamma\qex$ in the supremum (proof of Lemma~\ref{lem:split}). A long step then reaches the fixed point in fewer epochs, because each update carries the value of the end point over a long arc of the orbit. When the quadratic form is only a local Lyapunov function, as for the closed loops with an equilibrium of Appendix~\ref{app:bench}, $C_t$ can decrease along a trajectory. The fixed point then overestimates $\Phi$ for every step, so no step is exact there. In every case the method compares the two settings of Section~\ref{sec:train} on seed 0 by the share of box samples in the estimate, after level selection and the final test and before fine-tuning. It then keeps the setting with the larger share for all seeds. The comparison uses no reference labels, and both estimates have passed the same tests. The two settings differ in the cost scale and the exit weight as well as in the step, because each was developed as a whole (see the training setup above). Table~\ref{tab:step} in Appendix~\ref{app:bench} lists the comparison for every system.

\paragraph{Shaping cost.} When the inner layer holds at least one percent of the box, we add a shaping cost $\kappa\rho$ with $\kappa=0.3$ to $\mathcal L$, to steepen $v$ just outside the inner layer. Otherwise $\kappa=0$, as on the quadrotor. The quadrotor also uses the weight 10 for the finite-step penalty, against 100 on the unicycle, and no exponential loss weight during training. Here $\rho\in[0,1]$ is a smooth nondecreasing function of $\gamma\qex$, equal to zero below $c_I/2$ and to one above $c_I$, where $c_I$ is the level of the inner layer. Its integral is not known in closed form, so $\Phi$ absorbs it. With $C_t(x)=\kappa\int_0^t\rho(y(s;x))\,ds$ in physical time, the decomposition $V=\gamma\qex+\Phi$ holds with $\Phi(x)=\sup_{t\ge0}[C_t(x)+\psi(y(t;x))]$, under the conventions of Lemma~\ref{lem:split}, and \eqref{eq:dp} becomes
\begin{equation}\label{eq:dpk}
\Phi(x)=\max\Big\{\max_{0\le s\le\tau}\big[C_s(x)+\psi(y(s;x))\big],\ C_\tau(x)+\Phi(y(\tau;x))\Big\}.
\end{equation}
In normalized time the accumulated cost is $\bar C_s(x)=\int_0^s(\kappa\rho/a)(\bar y(s';x))\,ds'$. Training then stores $\bar C_\tau(x)$ with the end point, the running maximum is $R_\tau(x)=\max_{0\le s\le\tau}[\bar C_s(x)+\psi(\bar y(s;x))]$, and the target is $\mathcal U(x)=\exp(-\max\{R_\tau(x),\ \bar C_\tau(x)+\Phi_{\bar\theta}(x_\tau)\})$. The decrease penalty asks for $L_{f_{\mathrm{cl}}}\Phi_\theta\le-\kappa\rho$ and the obstacle penalty for $\Psi_\theta>\psi$ where $L_{f_{\mathrm{cl}}}\psi>-\kappa\rho$. For $\kappa=0$ all of these reduce to the expressions of Section~\ref{sec:method}. Theorem~\ref{thm:main} is unaffected, because $V_\theta\ge\gamma\qex$ still holds.

\paragraph{Penalties.} Lie derivatives in the decrease penalty and the obstacle penalty are taken at points of differentiability. The finite-step penalty acts on steps of unit length in $\Oad$ from $x$ to $x^{+}$. It asks for $V_\theta(x^{+})\le V_\theta(x)-\chi[\gamma(\qex(x)-\qex(x^{+}))+C_1]$ with $\chi=0.2$ and $C_1$ the accumulated shaping cost over the unit step, and it is applied to the values $e^{-V_\theta}$. An error in $V_\theta$ at a level $V$ moves $v_\theta$ by only the factor $e^{-V}$ of it, so without the exponential loss weight the loss would carry almost no gradient at the levels near one. Every five epochs, the $10^4$ states of a fresh pool of $2.5\times10^5$ whose unit step violates the decrease most, without the margin $\chi$, join the data. Each comes with four Gaussian neighbours, and the data are kept in a rolling buffer of $10^5$. The network starts close to $\Phi_\theta=\psi$, which lies below the exact solution. Where $\Phi=\psi$, any network with $\Psi_\theta\le\psi$ represents $\Phi$ exactly, so the network has to approximate $\Phi$ only where $\Phi>\psi$, on states whose trajectories later come closer to a constraint. Every term carries the exponential loss weight $e^{2\min(V_\theta,V_{\max})}$ with $V_{\max}=3$, and none during training on the quadrotor. Table~\ref{tab:config} gives every constant.

\paragraph{Screening and retest.} Screening uses $3\times5\cdot10^4$ box samples and the additional samples of Table~\ref{tab:config2}. The three best snapshots are retested on $5\times10^5$ fresh box samples, and the best one is kept with the lowest level it received.

\section{Additional Results}\label{app:extra}
\paragraph{Unsafe samples and the final test.} The direct test of unsafe states draws $10^6$ i.i.d.\ box samples per case and simulates each sample in the union of the four estimates until it reaches the inner layer. Of the 641\,790, 815\,296 and 135\,702 samples in the unions, 1, 1 and 1 leave the box first, and finer integration confirms all 3. These counts give the bounds of Section~\ref{sec:exp}, and each bound holds for every seed. By Theorem~\ref{thm:main}(iii), the annulus of every seed whose estimate contains an unsafe sample holds counterexamples that the final test missed. The final test found a counterexample and lowered the level in seven of the twelve runs (Table~\ref{tab:seeds}).

\begin{table}[t]
\centering\footnotesize\setlength{\tabcolsep}{3pt}
\caption{Per-seed results. Before: the trained network. Round: selected fine-tuning round (0 is the trained network). Sel.\ and Final: level after the retest and after the final test. Annulus: final-test samples between $c_1$ and the selected level. FP as in Table~\ref{tab:main}.}
\label{tab:seeds}
\begin{tabular}{llcccccccc}
\toprule
Case & Seed & $c_2$ before & Cov.\ before & Round & $c_2$ sel. & $c_2$ final & Annulus & Cov. & FP\\
\midrule
Circle & 0 & 0.8695 & 71.99 & 4 & 0.9700 & 0.9700 & 574520 & 74.04 & 0.000\\
Circle & 1 & 0.8570 & 69.16 & 4 & 0.9768 & 0.9768 & 544826 & 70.46 & 0.000\\
Circle & 2 & 0.8539 & 69.82 & 3 & 0.9786 & 0.9739 & 570552 & 72.52 & 0.000\\
Circle & 3 & 0.8358 & 69.23 & 4 & 0.9687 & 0.9687 & 548199 & 70.85 & 0.000\\
Sinusoid & 0 & 0.9071 & 82.53 & 1 & 0.9770 & 0.9770 & 784823 & 83.75 & 0.000\\
Sinusoid & 1 & 0.8668 & 81.64 & 1 & 0.9722 & 0.9700 & 790751 & 84.01 & 0.000\\
Sinusoid & 2 & 0.8777 & 80.96 & 2 & 0.9609 & 0.9609 & 772925 & 82.31 & 0.000\\
Sinusoid & 3 & 0.9099 & 83.16 & 4 & 0.9697 & 0.9677 & 787672 & 83.65 & 0.000\\
Quadrotor & 0 & 0.3140 & 24.86 & 0 & 0.3922 & 0.2506 & 151731 & 22.86 & 0.000\\
Quadrotor & 1 & 0.1226 & 13.53 & 0 & 0.2991 & 0.1226 & 135868 & 13.53 & 0.000\\
Quadrotor & 2 & 0.3431 & 23.73 & 0 & 0.3689 & 0.2472 & 136567 & 21.03 & 0.000\\
Quadrotor & 3 & 0.2541 & 23.15 & 0 & 0.3472 & 0.1823 & 145811 & 19.07 & 0.000\\
\bottomrule
\end{tabular}
\end{table}
\begin{table}[t]
\centering\small
\caption{Ablations on seed 0 before fine-tuning, with the same level selection and final test. FP: share (\%) of reference states inside the estimate and outside the simulated region of attraction.}
\label{tab:ablation}
\begin{tabular}{llccc}
\toprule
Case & Variant & $c_2$ & Coverage & FP\\
\midrule
Circle & full method & 0.869 & 72.0 & 0.000\\
Circle & PDE residual loss alone & 0.195 & 10.5 & 0.000\\
Circle & safety-value baseline & 0.743 & 11.1 & 0.002\\
Circle & no finite-step penalty, no counterexamples & 0.768 & 63.4 & 0.000\\
Circle & no exponential loss weight & 0.777 & 66.7 & 0.000\\
Circle & sum form $\Phi=\psi+\Psi_\theta$ & 0.995 & 9.9 & 0.000\\
Sinusoid & full method & 0.907 & 82.5 & 0.000\\
Sinusoid & safety-value baseline & 0.928 & 33.1 & 0.002\\
Quadrotor & full method & 0.314 & 24.9 & 0.000\\
Quadrotor & safety-value baseline & 0.483 & 1.3 & 0.000\\
\bottomrule
\end{tabular}
\end{table}

\paragraph{How reliable is the direct test of unsafe states?} It is at least as accurate as the reference simulation. The test uses seeds disjoint from all other tests, the RK4 integration of the normalized field with substeps of length $0.025$ as in the final test, and $t_{\max}=200$. Every sample that reached $\I$ did so within 8 units of normalized time. On 9000 reference states, the test and the reference simulation disagree on 41, and the reference simulation with a four times finer step agrees with the test on 38 of them. In the other 3, the test counts as unsafe a state that the finer reference finds converging, which is the conservative direction. None of the disagreeing states lies in an estimate, so this comparison does not test the safe labels inside the estimates. The 3 samples counted as unsafe still leave $\Oad$ with four and sixteen times finer substeps, and the reference simulation with step $0.001$ finds that they leave the box.

\paragraph{What do the two tests leave open?} With $N=10^6$ and $\varepsilon=\ln(10^6)/N=1.38\times10^{-5}$, Proposition~\ref{prop:scen} says with confidence $1-10^{-6}$ that the uniform measure of counterexamples in the annulus is at most $\varepsilon$. As noted in Section~\ref{sec:guarantees}, this alone does not bound the measure of states of the outer layer outside $\D(M)$. The reason is that a function can decrease everywhere except on a set of small measure while most trajectories diverge \citep{kenanian2019}. Both tests integrate numerically and check membership in $\Oad$ at substeps of length $0.025$, so a crossing of the singular set between substeps could go unnoticed. The test of Proposition~\ref{prop:unsafe} also checks the sign of $\det D$ at every substep, which detects a sign change of $\det D$ between substeps but not a crossing of the other constraints. The union of the outer layers over several seeds is therefore also tested on samples only. A deterministic check of the hypothesis of Theorem~\ref{thm:main}(iii) needs an enclosure of the flow over a unit step on every cell of the annulus, and we did not compute one.

\paragraph{What limits the outer layer?} Both the level and the fit limit it, and the fit matters most on the quadrotor. The exact value function cut at the level of the candidate covers 85.4\,\%, 88.1\,\% and 46.8\,\% of the simulated region of attraction, and the outer layers cover 84\,\%, 95\,\% and 41\,\% of these shares. On the unicycle the levels lie above 0.96, and the states that the exact value function leaves out at these levels have trajectories that pass near a constraint before converging. They therefore have a large $\Phi$, and only a level closer to one would include them. On the quadrotor, counterexamples hold the level between 0.12 and 0.25, while the exact value function cut at $0.9$ covers 80.5\,\%. At these levels the bound $V_\theta\ge\gamma\qex$ excludes 37.5\,\% of the simulated region of attraction, against 0.1\,\% on the circle and 0.0\,\% on the sinusoid.

\paragraph{What do the components contribute?} Fine-tuning helps on the unicycle and is never selected on the quadrotor, where screening kept the trained network in four of four runs. Each of these networks therefore has two final tests, one before and one after fine-tuning, and its reported level is the lower of the two (Table~\ref{tab:seeds}). The draw of the final test alone moves the coverage of a seed by up to 7.0 points. Table~\ref{tab:ablation} removes components on the circle, where the PDE residual loss alone, without fitted value iteration, penalties, counterexamples, exponential loss weight or weight averaging, reaches 10.5\,\%. The sum form replaces the maximum in \eqref{eq:ansatz} by $\Phi_\theta=\psi+\Psi_\theta$, and it has neither the obstacle penalty nor the one-sided loss of Section~\ref{sec:train}, because both refer to the maximum in \eqref{eq:ansatz}.

\paragraph{Does the integration step change the reference labels?} It changes few of them, at most 0.74\,\% on a case. We simulated $5000$ of the $5\times10^4$ reference states of each case again with an RK4 step of $0.0025$ in place of $0.01$. The label changes on 0.24\,\%, 0.24\,\% and 0.74\,\% of these states on the circle, the sinusoid and the quadrotor.

\paragraph{What limits the inner layer?} Contact with the constraints bounds it from above, and the interval verification stops below that bound. On samples, the sublevel sets of $\gamma\qex$ first meet a regularity constraint, or a face of the box where the field points outward, at levels of at most 0.0417, 0.0188 and 0.0019 on the three cases. The verified levels are 0.0350, 0.0170 and 0.0002 (Table~\ref{tab:inner}), and on the quadrotor the verification at the next level exceeds its budget of cells. The first contact lies near the centre of the circle, on the speed constraint for the sinusoid, and on the zero-thrust face for the quadrotor. Thrust enters $z$ only through the acceleration coordinates, so a state with no thrust can have a small $\qex$, and the inner layer of the quadrotor is negligible.

\paragraph{Computation time.} All computations run on the CPU of a laptop (Apple M1). The trainings ran eight at a time on one thread each, next to the level selection of other runs, and took 44 to 73 minutes per seed under this load. Level selection took 1 to 13 minutes, fine-tuning 2 to 59 minutes, and the interval verification 110 to 3626 seconds per case.

\end{document}